\pdfoutput=1

\documentclass[11pt]{article}

\usepackage[preprint]{acl}

\usepackage{times}
\usepackage{latexsym}

\usepackage[T1]{fontenc}

\usepackage[utf8]{inputenc}

\usepackage{microtype}

\usepackage{inconsolata}

\usepackage{graphicx}

\usepackage{tabularx}
\usepackage{float}
\usepackage{arabtex}
\usepackage{utf8}
\setcode{utf8}

\usepackage{longtable}
\usepackage{enumitem}
\usepackage{amsmath}
\usepackage{multirow}
\usepackage{booktabs}
\usepackage{listings}
\usepackage{xcolor}
\colorlet{punct}{red!60!black}
\definecolor{background}{HTML}{EEEEEE}
\definecolor{delim}{RGB}{20,105,176}
\colorlet{numb}{magenta!60!black}

\lstdefinelanguage{json}{
    basicstyle=\ttfamily\fontfamily{fvm}\selectfont,
    showstringspaces=false,
    breaklines=true,
    frame=lines,
    backgroundcolor=\color{background},
    literate=
     *{0}{{{\color{numb}0}}}{1}
      {1}{{{\color{numb}1}}}{1}
      {2}{{{\color{numb}2}}}{1}
      {3}{{{\color{numb}3}}}{1}
      {4}{{{\color{numb}4}}}{1}
      {5}{{{\color{numb}5}}}{1}
      {6}{{{\color{numb}6}}}{1}
      {7}{{{\color{numb}7}}}{1}
      {8}{{{\color{numb}8}}}{1}
      {9}{{{\color{numb}9}}}{1}
      {:}{{{\color{punct}{:}}}}{1}
      {,}{{{\color{punct}{,}}}}{1}
      {\{}{{{\color{delim}{\{}}}}{1}
      {\}}{{{\color{delim}{\}}}}}{1}
      {[}{{{\color{delim}{[}}}}{1}
      {]}{{{\color{delim}{]}}}}{1},
}
\title{Nile-Chat: Egyptian Language Models for Arabic and Latin Scripts}

\author{
 \textbf{Guokan Shang\textsuperscript{1}$^\star$$^\dagger$},
 \textbf{Hadi Abdine\textsuperscript{1}$^\star$},
 \textbf{Ahmad Chamma\textsuperscript{1}$^\star$},
 \textbf{Amr Mohamed\textsuperscript{1}$^\star$},\\
 \textbf{Mohamed Anwar\textsuperscript{1}},
 \textbf{Abdelaziz Bounhar\textsuperscript{1}},
 \textbf{Omar El Herraoui\textsuperscript{1}}, \\
 \textbf{Preslav Nakov\textsuperscript{1}},
 \textbf{Michalis Vazirgiannis\textsuperscript{1,2}$^\dagger$},
 \textbf{Eric Xing\textsuperscript{1}}
\\
\\
 \textsuperscript{1}MBZUAI,
 \textsuperscript{2}Ecole Polytechnique
\\
 \small{
   $^\dagger$Correspondence: \texttt{\{guokan.shang, michalis.vazirgiannis\}@mbzuai.ac.ae}
 }
}

\begin{document}
\maketitle
\def\thefootnote{$\star$}\footnotetext{These authors contributed equally.}\def\thefootnote{\arabic{footnote}}

\begin{abstract}
We introduce \texttt{Nile-Chat-4B}, \texttt{3x4B-A6B}, and \texttt{12B}\footnote{\url{https://hf.co/MBZUAI-Paris/Nile-Chat-12B}}, a collection of LLMs for Egyptian dialect, uniquely designed to understand and generate texts written in both Arabic and Latin scripts.
Specifically, with \texttt{Nile-Chat-3x4B-A6B}, we introduce a novel language adaptation approach by leveraging the Branch-Train-MiX strategy to merge script-specialized experts, into a single MoE model.
Our Nile-Chat models significantly outperform leading multilingual and Arabic LLMs—such as LLaMa, Jais, and ALLaM—on our newly introduced Egyptian evaluation benchmarks, which span both understanding and generative tasks. Notably, our \texttt{12B} model yields a 14.4\% performance gain over \texttt{Qwen2.5-14B-Instruct} on Latin-script benchmarks.
All our resources are publicly available. We believe this work presents a comprehensive methodology for adapting LLMs to dual-script languages, addressing an often overlooked aspect in modern LLM development.
\end{list}
\end{abstract}

\section{Introduction}

Egyptian Arabic (also known as \textit{Masri}) is the most widely spoken variety of Arabic, with over 100 million native speakers in Egypt and broader mutual intelligibility across the Arab world\footnote{\url{https://en.wikipedia.org/wiki/Egyptian_Arabic}}.
It differs substantially from Modern Standard Arabic (MSA) in phonology, vocabulary, and grammar.
A notable feature of this dialect is its widespread dual-script usage: native speakers often write Egyptian Arabic in both Arabic script and a Latin-based script commonly referred to as \textit{Arabizi} or \textit{Franco-Arabic} (e.g., “7aga gameda” for \<حاجة جامدة>). 

Despite the pervasiveness of this dual-script setting, most Large Language Models (LLMs) for Arabic fail to support it adequately. Existing models either focus on MSA or partially support dialects, and none are trained to handle the Latin script. Moreover, no prior LLMs have \textit{explicitly} targeted a single language across two scripts.

\begin{figure*}[t]
    \centering
    \includegraphics[width=\textwidth]{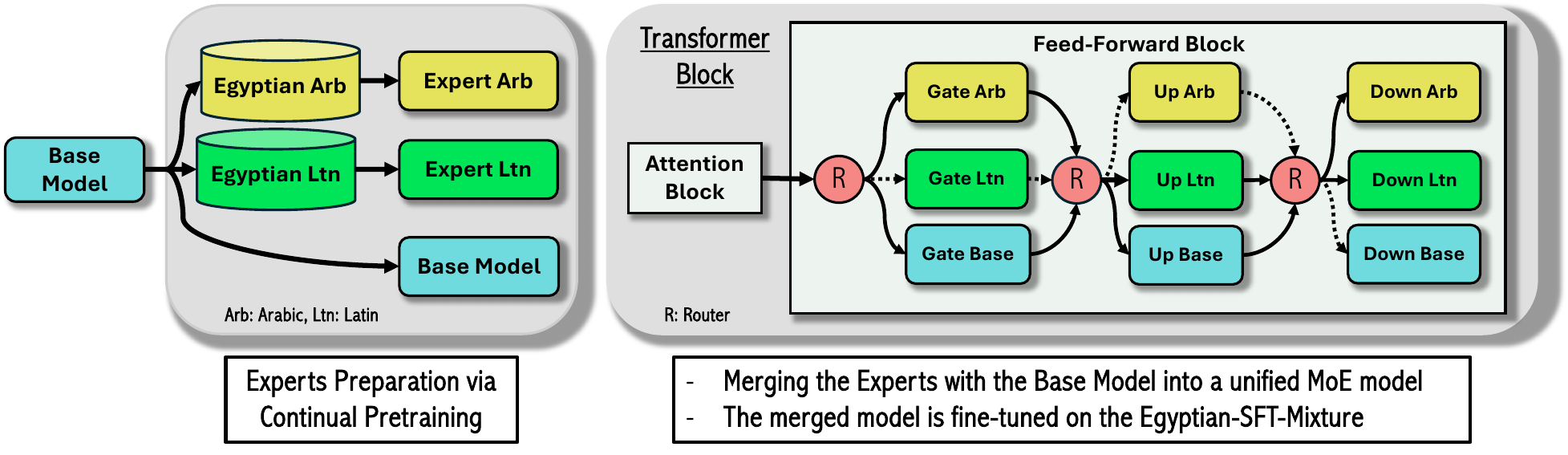} 
    \caption{The training of \texttt{Nile-Chat-3x4B-A6B} using the \textit{Branch-Train-MiX} (BTX) strategy. \textbf{Left}: Two experts are first \textit{continual pre-trained} on Arabic-script and Latin-script corpora, respectively. \textbf{Right}: A \textit{Top-2} token routing example within a transformer block, where the two script-specialized Experts have been merged with the Base Model into a unified \textit{Mixture-of-Experts} (MoE) model through \textit{instruction-tuning}.}
    \label{fig:btx-nile} 
\end{figure*}

We introduce Nile-Chat\footnote{We chose \emph{Nile} to reflect the cultural and geographical significance of the Nile river, which traverses Egypt.}, an LLM family for Egyptian Arabic that natively supports two scripts. We release three model variants: dense models in \texttt{4B} and \texttt{12B}, and \texttt{Nile-Chat-3x4B-A6B}: a \textit{Mixture-of-Experts} (MoE) model trained using the \textit{Branch-Train-MiX} (BTX) method \citep{sukhbaatar2024branch}. As shown in Figure \ref{fig:btx-nile}, we merge script-specialized experts, each trained on either Arabic-script or Latin-script Egyptian data, into a unified MoE that dynamically routes tokens to the appropriate expert. This modular approach enables scalable adaptation without sacrificing performance or efficiency.

All Nile-Chat models undergo a full training pipeline with dual-script data we created---including continual pre-training on Egyptian Arabic corpora (e.g., transcripts, forum posts, and song lyrics), followed by fine-tuning on a variety of instruction tasks, and a final alignment-tuning stage for safety and preference adjustment.
To support the evaluation, we also introduce a comprehensive evaluation suite covering both understanding (e.g., MMLU, HellaSwag) and generation (e.g.,~translation, transliteration) tasks in Arabic and Latin scripts. 
Nile-Chat models consistently outperform competitive baselines including LLaMa, ALLaM, Jais, and Qwen2.5 across all Egyptian-specific benchmarks. Notably, our \texttt{12B} model improves Latin-script benchmark performance by 14.4\% over \texttt{Qwen2.5-14B-Instruct}.

To the best of our knowledge, Nile-Chat is the first LLM to provide script-aware support for a widely spoken dialect. All models, data, and evaluation code are released publicly. We hope that our work will inspire further research on LLMs for underrepresented and dual-script languages.

\section{Related Work}
\paragraph{Arabic LLMs and Dialectal Models.}
 The proliferation of Arabic-specific LLMs has included models like Jais \citep{sengupta2023jais}, AceGPT \citep{huang-etal-2024-acegpt}, and ALLaM \citep{bari2024allam}, trained primarily on MSA and English, often overlooking dialects. More closely related to our work, Atlas-Chat \citep{shang-etal-2025-atlas} introduced LLMs for Moroccan Arabic, demonstrating that dialectal models can outperform general multilingual models. Our Nile-Chat advances this paradigm, explicitly supporting the widely used Egyptian dialect, and uniquely, as written in both Arabic and Latin scripts.

\smallskip

\noindent\textbf{Romanized Arabic and Dual-script Languages.}
Romanized Arabic---also known as \emph{Arabizi} or \emph{Franco-Arabic}---is widely used in informal communication, especially among youth~\citep{yaghan2008arabizi, alghamdi2018arabizi}. It transcribes Arabic words using Latin characters and numerals (e.g., ``3'' for \<ع>) and remains common in digital communication, despite broad support for Arabic script.

Prior work has focused on \emph{detecting} and \emph{transliterating} Arabizi into Arabic script~\citep{darwish2013arabizi}, treating it as a noisy input to be normalized. In contrast, we treat both scripts as \emph{native inputs and outputs}, allowing the model to directly understand and generate Egyptian Arabic in either form.

Other languages such as Hindi, Serbian, and Kazakh \citep{koto2025llama} also exhibit dual-script usage. In Hindi, for example, the Nanda model~\citep{choudhury2025llama} enhances robustness to Latin-script Hindi by augmenting the training data. Our work goes one step further; we use script-specialized experts within an MoE architecture to model each script \textit{explicitly}. To the best of our knowledge, Nile-Chat is the first LLM for Arabic that supports both native and Latin scripts in a unified framework.

\smallskip

\noindent\textbf{Mixture-of-Experts.}
MoE models \citep{jiang2024mixtral} efficiently scale LLM capabilities by selectively activating sub-networks. The recent Branch-Train-Mix (BTX) strategy \citep{sukhbaatar2024branch} allows fine-grained merging of specialized expert models, significantly reducing training costs. Our \texttt{Nile-Chat-3x4B-A6B} model innovatively applies BTX to script-specialized experts, efficiently integrating expertise in both Arabic and Latin scripts within a single model. This novel strategy demonstrates the viability of MoE architectures for linguistic specialization.

\section{Dual-Script Training Data}
The datasets feeding the Nile-Chat training fall into three broad categories:

\noindent\textbf{Continual Pre-training}: large‐scale unlabeled Egyptian Arabic text drawn from audio / video transcripts, online forums, song lyrics, Wikipedia dumps, and web‐scale crawls (see \S\ref{sec:pre-training_datasets}).\\
\textbf{Instruction-tuning}: prompt–response pairs covering a variety
of instruction tasks, assembled from native Egyptian sources, and high-quality English translations (see \S\ref{sec:instruction_datasets}).\\
\textbf{Alignment-tuning}: preference pairs used with Direct Preference Optimization to refine safety and mitigate undesirable behavior (see \S\ref{sec:alignment_datasets}).

Across all of the above datasets, we ensure that roughly 25\% is represented in the Latin script, complementing the Arabic-script majority and reflecting real-world usage patterns. 
The remainder of this section details each category in turn.

\subsection{Continual Pre-training Datasets}
\label{sec:pre-training_datasets}

As Egyptian Arabic is primarily used in spoken form, we first curated 854K audio / video \textbf{transcripts} to better capture its natural usage, yielding a total of 829M words.
To broaden coverage, we supplemented the collection with publicly available datasets spanning diverse domains and styles.
These include the \textbf{EFC-mini} (Egyptian Forums Corpus-mini)~\citep{qarah2024egybert}, the \textbf{EDC} (Egyptian Datasets Collection)\footnote{\url{https://github.com/Mostafanofal453/2.5-Million-Rows-Egyptian-Datasets-Collection}}, the \textbf{Egyptian Wikipedia dump}\footnote{\url{https://dumps.wikimedia.org/arzwiki/}}, the Egyptian subset of the \textbf{ADD} (Arabic Dialects Dataset)\footnote{\url{https://elhaj.uk/corpora.html}}, the Egyptian partition of \textbf{FineWeb-2}~\citep{penedo2025fineweb2}, the Egyptian subset of the \textbf{Habibi lyrics corpus}~\citep{el-haj-2020-habibi}, and a small collection of scraped forum posts from \textbf{Fatakat}\footnote{\url{https://forums.fatakat.net}}. Full details are provided in Appendix~\ref{appendix:pre-training_datasets}.

The resulting pre-training corpus contains 1.15B words, predominantly in Arabic script. To balance this, we used Claude to transliterate a portion into Latin script (see the prompt in Appendix~\ref{sec:pre-training_transliteration}).
For this, we selected samples from the transcripts, EFC-mini, and EDC datasets, which feature informal content such as conversations, social media posts, and user comments—domains where Latin script is frequently used in practice.
This process resulted in a total of 255M words in Latin script.

\subsection{Instruction-tuning Datasets}
\label{sec:instruction_datasets}

\begin{figure}[t]
    \centering
    \includegraphics[width=0.52\textwidth]{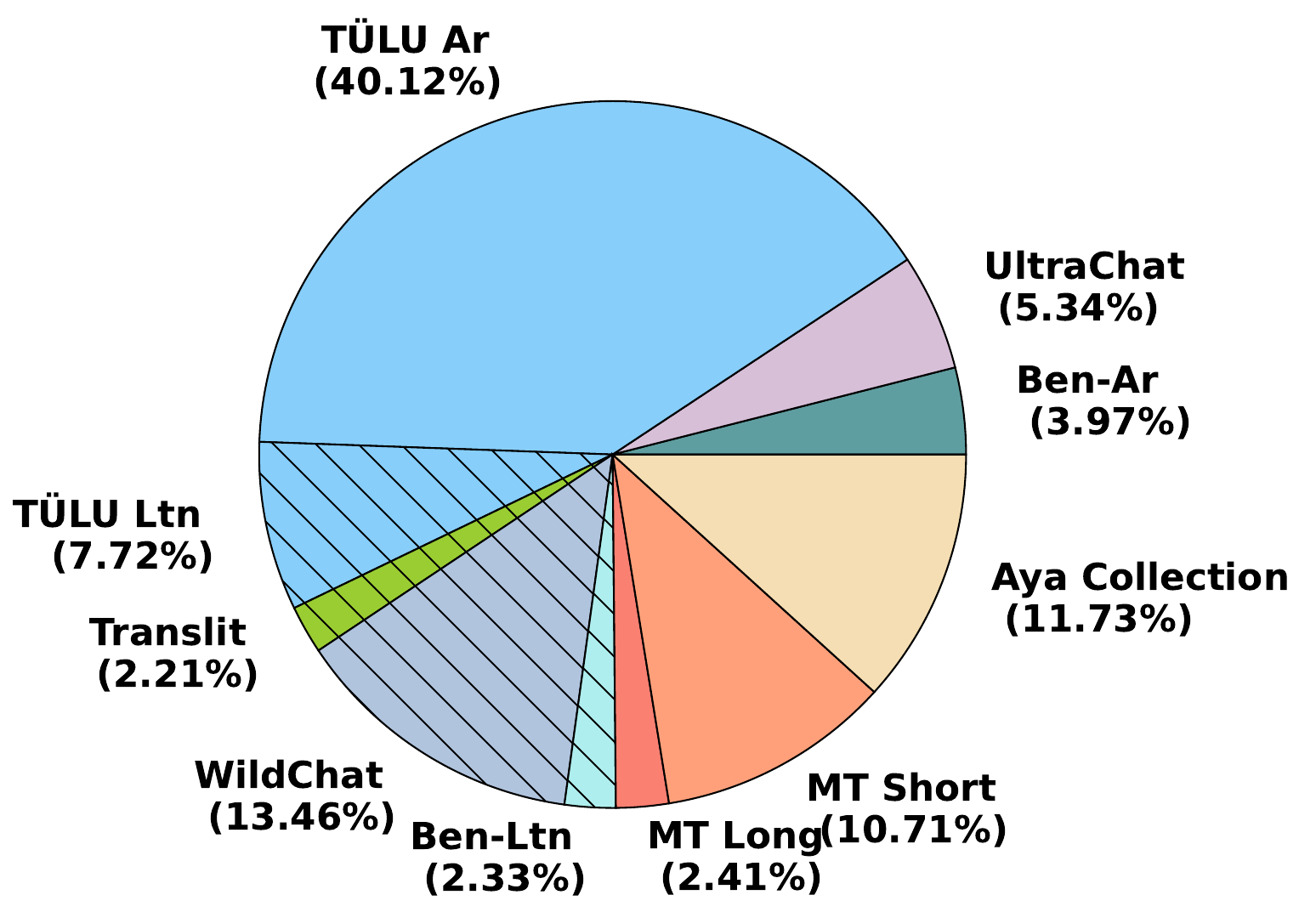} 
    \caption{Composition of our Egyptian-SFT-Mixture instruction-tuning dataset. The acronyms "MT", "Ar", "Ltn", "Translit" and "Ben" are used to denote "Machine Translation", "Arabic", "Latin", "Transliteration", and "Benchmarks Training Set" respectively. The hatched regions represent parts in Latin script.}
    \label{fig:pie_chart} 
\end{figure}

To fine-tune the models for instruction following in Egyptian Arabic, we created the \textbf{Egyptian-SFT-Mixture}\footnote{\url{https://hf.co/datasets/MBZUAI-Paris/Egyptian-SFT-Mixture}} of 1.85M instructions by consolidating multiple sources, as illustrated in Figure~\ref{fig:pie_chart}. We began by incorporating publicly available datasets from prior work. To broaden coverage across domains and tasks, we translated some English instruction datasets into Egyptian Arabic. Finally, we augmented the mixture with data for translation between Egyptian, English, and MSA, as well as transliteration where users request conversion between Arabic and Latin scripts.
The dataset is formatted as user-assistant messages in Appendix~\ref{app:tulu:format}.

\subsubsection{Existing Egyptian Instruction Datasets}
\label{sec:data:aya_collection}
To the best of our knowledge, the \textbf{Aya Collection}~\citep{singh-etal-2024-aya} is the only large-scale multilingual instruction dataset that provides a readily usable subset in Egyptian Arabic, with over 3.5M samples across a wide range of tasks. These include \textit{paragraph writing}, \textit{text classification}, \textit{paraphrase identification}, \textit{question-answering}, \textit{summarization}, and \textit{text simplification}.
To ensure language consistency, we applied a \textit{Glotlid}-based language identification filter~\citep{kargaran-etal-2023-glotlid} to exclude non-Egyptian Arabic samples.

\subsubsection{Translated English Instruction Datasets}
\label{sec:data:english_tulu}

We began by examining instruction-tuning datasets used to fine-tune recent state-of-the-art models.
\textbf{TÜLU Collection} stands out for its broad domain coverage, including \textit{instruction following}, \textit{knowledge recall}, \textit{reasoning}, and \textit{safety}. The dataset mixture was systematically designed based on findings from ablation studies of both human-annotated and AI-generated data, with a deliberate emphasis on complexity and diversity. Appendix \ref{app:tulu:composition} presents descriptions of each of the nested datasets, and describes how the subset was sampled. TÜLU-v3-mix \citep{lambert2024t} is the successor of TÜLU-v2-mix \citep{ivison2023camels} with some intersected samples. We chose in this work to include both versions after eliminating the nested datasets where a newer version is provided, and performed a string-based de-duplication step for the remaining parts where 9,660 samples were removed. This forms our initial TÜLU-v2\&3-mix dataset.

To improve quality, we first applied a preliminary filtering process to the v2\&3 dataset, removing instructions that were unsuitable for typical Egyptian users or likely to lose meaning in translation—such as \textit{scientific content}, \textit{translation tasks}, and \textit{non-English text}.
For English-to-Egyptian Arabic translation, we compared \texttt{GPT-4o} and \texttt{Claude 3.5 Sonnet}. Based on qualitative evaluation, Claude produced more natural and dialect-appropriate outputs, and was ultimately selected for translating the remaining data.
Finally, to rectify the issues introduced by the automatic translation, a series of post-processing measures were implemented. All details are provided in Appendix \ref{app:tulu:translation}.
Similar to our pre-training data, we selected a subset of the data—primarily focused on chat-style examples across various topics—and processed them into the Latin script.

Although TÜLU-v2\&3-mix includes instructions from diverse domains, it contains only around 38K multi-turn conversations (with at least two turns). To improve the model's ability to sustain longer dialogues~\citep{zhao2024long}, we incorporated data from \textbf{UltraChat}~\citep{ding2023enhancing}, a multi-round dialogue dataset that covers world knowledge, writing, and creative tasks. The dataset contains over 300K conversations, each with a minimum of five exchanges. We selected the longest examples—those with 7 to 8 turns—and applied the same processing procedures described for v2\&3.

To further increase Latin-script representation, we also included data from \textbf{WildChat}~\citep{zhao2024wildchat}, a dataset of 1M dialogues between users and ChatGPT, organized by script, language, and country. From the English subset (over 450K samples), we selected the first 300K conversations sorted by ascending length—based on the assumption that the Latin script is more common in short- to mid-length exchanges. These samples were translated into Egyptian Arabic in the Latin script and post-processed following the same procedure described above.

\subsubsection{Translation and Transliteration Tasks}
The final portion of our instruction data specifically targets two tasks: translation and transliteration.

\smallskip

\noindent\textbf{\textit{Short Sentence Translation}}

We incorporated four publicly available translation datasets into our mixture. These include \textbf{EGY\_MSA\_Translation}~\citep{faheem2024improving}, a parallel corpus of Egyptian Arabic and MSA sentences collected from social media; \textbf{ArzEn-MultiGenre}~\citep{al2024arzen}, which includes professionally translated texts across songs, novels, and TV subtitles; \textbf{Egyption\_2\_English}\footnote{\url{https://hf.co/datasets/Abdalrahmankamel/Egyption_2_English}}, a 22k-sample dataset of everyday bilingual sentences; and \textbf{Oasst2-9k-translation}\footnote{\url{https://hf.co/datasets/ahmedsamirio/oasst2-9k-translation}}, which provides English prompts aligned with Egyptian Arabic and MSA outputs, generated using \texttt{GPT-4o}. Detailed descriptions are provided in Appendix \ref{appendix:instruction-tuning_datasets}.

The collected samples were converted into training instructions using randomly selected Egyptian-based templates (see Appendix~\ref{app:instruction_templates:machine_translation}). We cover four translation directions: Egyptian Arabic to English, to MSA, and vice versa. To enhance multi-turn translation capabilities, a portion of the dataset includes 3-shot examples and 3-turn conversations. 10\% of the data is reserved for evaluation.

\smallskip

\noindent\textbf{\textit{Long Document Translation}}

The above collection of translation samples---whether derived from native translators or advanced models---mostly consists of short sentences. To equip the model with the ability to handle mid- to long-form translation (i.e., multi-line documents), we further used data from the Egyptian Wikipedia dump. We removed entries that were not relevant for translation, such as indicators of missing content, empty pages, and astronomy-related topics, which are overrepresented in the dump. We retained documents with word counts between 90 and 1,500 and applied a \textit{Glotlid} filter to eliminate non-Egyptian Arabic samples. These documents were then translated into English and MSA using Claude, and subsequently transformed into training instructions using the template provided in Appendix~\ref{app:instruction_templates:machine_translation}.

\smallskip

\noindent\textbf{\textit{Transliteration}}

To enable our model to perform script conversion between the Arabic and the Latin scripts, we use the Egyptian Forums Corpus (EFC), introduced by \citet{qarah2024egybert}, which contains user-generated texts from various Egyptian online forums. To promote sample diversity, we removed frequent keyterms related to sports. We then selected sentences with lengths between 50 and 70 words and applied a \textit{Glotlid} language filter to ensure dialectal consistency. From the filtered set, we retained final samples and converted them from the Arabic to the Latin script to build a parallel corpus. These were then transformed into training instructions using the templates given in Appendix~\ref{app:instruction_templates:transliteration}.

\subsection{Alignment-tuning Datasets}
\label{sec:alignment_datasets}

To improve the overall model behavior, we applied a targeted alignment phase using Direct Preference Optimization (DPO)~\cite{rafailov2023direct}, combining on- and off-policy strategies. This was motivated by human evaluations of our SFT-stage model trained only on our pre-training and instruction data, which revealed several issues, including:

\smallskip

\noindent\textbf{Overly Cautious.}
We observed that the SFT-stage model frequently refused to answer legitimate questions due to excessive caution. To address this, we leveraged 50\% of the safety-related instructions retained from the SFT phase. For these samples, we applied an on-policy DPO strategy: the original assistant output was treated as the preferred response, while a corresponding rejected response was generated using the SFT-stage model itself.

\smallskip

\noindent\textbf{Excessive Code-Switching.}
We observed that the SFT-stage model exhibited excessive code-switching between Arabic and English \citep{mohamed2025lost}, even when the prompt was exclusively written in Arabic. To mitigate this behavior, we applied an off-policy correction procedure wherein instances from the SFT dataset exhibiting the identified patterns were selected and reformulated using Claude to produce more natural code-switched alternatives. The selection criteria and the correction prompt are described in detail in Appendix \ref{app:off-policy}.

\smallskip

\noindent\textbf{Failures in Instruction Tasks.}
Additionally, the SFT-stage model displayed shortcomings in several instruction-following capabilities, notably:
\begin{itemize}
\item  \textit{Length control}: The model frequently ignored explicit length requirements (e.g., producing a 400-word script when 600 words were requested).
\item  \textit{Stylistic control}: Rewriting or rephrasing with a specific tone (e.g., formal, humorous) was often inaccurate or superficial.
\end{itemize}
To address these issues, we again applied on-policy DPO strategy. 
We synthetically curated 1,000 minimal yet precise prompts, annotated poor completions from the SFT-stage model as rejections, and synthetically constructed new completions as positive demonstrations using Claude. 
The resulting preference pairs improved the model’s resilience to diverse user requests and yielded finer-grained control over its responses.
Our DPO datasets are publicly available\footnote{\url{https://hf.co/datasets/MBZUAI-Paris/Egyptian-DPO-OffPolicy}}.

\section{Training Details: Dense Models}
\label{nile:dense}

This section details the training setup across the pre-training, instruction-tuning, and alignment-tuning phases of \texttt{Nile-Chat-4B} and \texttt{12B} dense models.

\smallskip

\noindent\textbf{Base Model Selection.}
We adopt the base Gemma-3~\cite{team2025gemma} models as the starting point for training Nile-Chat, due to their superior performance on Arabic tasks in our preliminary evaluation compared to other state-of-the-art multilingual and Arabic-specialized models.

\smallskip

\noindent\textbf{Training Pipeline and Hyperparameters.} For our dense models, we merged all Arabic- and Latin-script datasets into a single corpus and trained on this unified mixture, in contrast to our MoE models described in Section \ref{sec:training_moe}.

\smallskip

\noindent\textit{\textbf{Continual Pre-training:}} We used Low-Rank Adaptation (LoRA) with rank \texttt{256} and alpha \texttt{128}. The optimizer is AdamW with $\beta_1 = \texttt{0.9}$ and $\beta_2 = \texttt{0.95}$. This stage is divided into:
\begin{itemize}
\item \textit{Continual pre-training.} We run the training for 1 epoch using our data from Section \ref{sec:pre-training_datasets}, with a learning rate of \texttt{8e-6}, a warmup ratio of \texttt{1\%}, and a cosine decay to \texttt{1e-6}.
\item \textit{Annealing phase}. During this phase, training gradually shifts focus to a smaller set of high-quality Egyptian Arabic data. 
We run the training for 1 epoch and set the learning rate to \texttt{3e-4} for the \texttt{4B} model and \texttt{5e-5} for the \texttt{12B} model, and a cosine decay to 0.
\end{itemize}

\smallskip

\noindent\textit{\textbf{Instruction-tuning (SFT):}} Next, we fine-tuned the model on our data from Section~\ref{sec:instruction_datasets}. We used LoRA with rank \texttt{256}
and alpha \texttt{128}.  We ran the training for 2 epochs,
and set the learning rate to \texttt{3e-5} for the 4B model and \texttt{2e-5} for the 12B model with a warmup ratio of \texttt{3\%}, linear decay to 0, and total effective batch size of \texttt{128}. The
loss is computed on the responses only. We used the AdamW optimizer with $\beta_{1} = \texttt{0.9}$ and $\beta_2 = \texttt{0.999}$.

\smallskip

\noindent\textit{\textbf{Alignment-tuning (DPO):}} Finally, we applied DPO to improve the overall model behavior, using the data constructed in Section~\ref{sec:alignment_datasets}. 
We followed standard DPO heuristics, notably reducing the SFT learning rate by an order of magnitude. Specifically, we evaluated learning rates of \texttt{3e-6} and \texttt{5e-6} with preference temperatures $\beta \in \{0.1, 0.5\}$, and compared full fine-tuning to LoRA. The experiments on the \texttt{Nile-Chat-4B} model showed that full fine-tuning with \texttt{3e-6} and $\beta = 0.5$ consistently performed better, both in benchmarks and human tests.  We adopted this configuration for the final alignment phase of the \texttt{Nile-Chat-12B} model.

We performed the training on 8×NVIDIA A100 80GB GPUs using Fully Sharded Data Parallel (FSDP) on AWS SageMaker. The maximum input context length was configured to \texttt{2,048} tokens.
We used \texttt{bfloat16} for faster training.

\section{Training Details: MoE Models} \label{sec:training_moe}

Recent literature has highlighted that dense models are prone to catastrophic forgetting—particularly during fine-tuning—as new inputs often overwrite previously acquired knowledge~\cite{li2024theory}. This effect is linked to data saturation, where model capacity is insufficient to retain all learned information. While scaling up dense models can alleviate forgetting to some extent, it comes at the cost of significantly higher inference budgets, since all parameters are used for every input. Mixture-of-Experts (MoE) models~\cite{lo2024closer} offer a more efficient alternative. By assigning tasks to specialized experts and routing at the token level~\cite{jiang2024mixtral}, MoEs isolate parameter updates, thereby reducing interference and preserving prior knowledge. This modular design enables MoEs to mitigate forgetting more effectively than dense models, while maintaining lower computational overhead.

Instead of training an MoE model from scratch, \citet{sukhbaatar2024branch} show a recycling strategy called \textbf{Branch-Train-Mix (BTX)}. This method constructs an MoE model by merging several pre-trained base models. Specifically, the feed-forward layers of these models are repurposed as distinct experts within a new MoE layer, while a trainable routing network assigns each token to the most relevant expert path. The remaining layers—such as attention and embeddings—are merged by averaging their parameters across the base models, forming a shared backbone. Finally, the resulting MoE model is fine-tuned on an SFT dataset to align the components and optimize joint performance.

As illustrated in Figure~\ref{fig:btx-nile}, we propose a novel LLM adaptation strategy for dual-script languages by applying BTX to script-specialized experts.
First, the base model is continually pre-trained on Arabic-script and Latin-script datasets separately to create script-specialized experts—differing from the unified training used for our dense models described in Section~\ref{nile:dense}. 
Second, the pre-trained experts and the base model are merged using the BTX scheme described above, resulting in a new MoE model with three experts—two of which are active per input—with a total of 6B activated parameters. This yields our final \textbf{\texttt{Nile-Chat-3x4B-A6B}} model. For comparison, we also merged the two script-specialized experts without including the base model, producing a \textbf{\texttt{2x4B-A6B}} variant.
We consider the three-expert variant as our primary model, as incorporating the base model as an additional expert integrates broader general knowledge and English capabilities that go beyond the scope of the script-specialized experts.
The unified MoE models then undergo two training phases: (1) \textit{SFT} using a LoRA setup with an alpha of \texttt{512}, a learning rate of \texttt{1e-4}, and an effective batch size of \texttt{256}. Since the English-centric base model is included as a third expert, we also mixed in Egyptian-SFT-Mixture a small amount of English instructions to recover its original English performance. (2) \textit{DPO} serves as the final alignment stage.

\begin{table*}[t]
\centering
\renewcommand{\arraystretch}{1.2}
\setlength{\tabcolsep}{4pt}
\large
\resizebox{\textwidth}{!}{
\begin{tabular}{p{6cm}cccccccc|c|ccc|ccc|ccc}
\toprule
\textbf{\normalsize Model} &
\multirow{2}{*}{\shortstack{\textbf{\small Egyptian} \\ \textbf{\small MMLU}}} &
\textbf{\small Belebele\_Arz} &
\multirow{2}{*}{\shortstack{\textbf{\small Egyptian} \\ \textbf{\small HellaSwag}}} &
\multirow{2}{*}{\shortstack{\textbf{\small Egyptian} \\ \textbf{\small PIQA}}} &
\multirow{2}{*}{\shortstack{\textbf{\small Egyptian} \\ \textbf{\small WinoGrande}}} &
\multirow{2}{*}{\shortstack{\textbf{\small Egyptian} \\ \textbf{\small OpenBookQA}}} &
\multirow{2}{*}{\shortstack{\textbf{\small Egyptian} \\ \textbf{\small RACE-H}}} &
\multirow{2}{*}{\shortstack{\textbf{\small Egyptian} \\ \textbf{\small RACE-M}}} &
\multirow{2}{*}{\shortstack{\textbf{\small Egyptian} \\ \textbf{\small AlpacaEval}}} &
\multicolumn{3}{c|}{\textbf{\small Long Translation}} &
\multicolumn{3}{c|}{\textbf{\small Short Translation}} &
\multicolumn{3}{c}{\textbf{\small Transliteration}} \\
\cmidrule(lr){11-13} \cmidrule(lr){14-16} \cmidrule(lr){17-19}
& & & & & & & & & &
\textbf{\small BLEU} & \textbf{\small chrF} & \textbf{\small BERTScore} &
\textbf{\small BLEU} & \textbf{\small chrF} & \textbf{\small BERTScore} &
\textbf{\small BLEU} & \textbf{\small chrF} & \textbf{\small BERTScore} \\
\midrule
\normalsize \textbf{gemma-3-4b-it} & 46.08 & 38.56 & 42.56 & 60.32 & 56.49 & 35.79 & 33.68 & 40.06 & 85.30 & 20.67 & 44.75 & 73.03 & 4.76 & 31.15 & 52.98 & 1.44 & 20.36 & 47.54 \\
\normalsize \textbf{jais-family-6p7b-chat} & 42.60 & 57.33 & 49.18 & 62.23 & 57.04 & 33.33 & 34.72 & 37.50 & 45.86 & 12.71 & 36.53 & 68.07 & 8.73 & 31.52 & 56.78 & 0.70 & 10.64 & 42.51 \\
\normalsize \textbf{jais-adapted-7b-chat} & 40.96 & 55.67 & 40.85 & 56.50 & 54.35 & 32.89 & 34.62 & 42.33 & 21.45 & 10.61 & 27.56 & 63.48 & 9.19 & 24.85 & 53.52 & 1.11 & 6.14 & 40.45 \\
\normalsize \textbf{Qwen2.5-7B-Instruct} & 45.74 & 64.22 & 45.47 & 58.02 & 56.41 & 38.70 & 35.45 & 41.76 & 58.80 & 19.89 & 44.80 & 73.64 & 11.34 & 36.31 & 54.96 & 2.74 & 20.63 & 49.32 \\
\normalsize \textbf{ALLaM-7B-Instruct-preview} & 60.08 & 67.67 & 57.29 & 66.10 & 62.18 & 40.04 & 39.50 & 45.17 & 69.55 & 26.57 & 52.59 & 78.34 & 25.20 & 48.12 & 65.97 & 2.10 & 18.92 & 49.42 \\
\normalsize \textbf{c4ai-command-r7b-arabic-02-2025} & 50.97 & 70.67 & 50.39 & 61.84 & 57.20 & 36.91 & 41.89 & 46.02 & 73.36 & 25.18 & 50.26 & 77.97 & 23.30 & 45.34 & 65.20 & 3.52 & 24.57 & 50.49 \\
\normalsize \textbf{Llama-3.1-8B-Instruct} & 42.88 & 55.89 & 43.10 & 57.97 & 54.27 & 35.57 & 34.41 & 40.34 & 52.35 & 12.90 & 32.58 & 68.76 & 9.06 & 28.56 & 54.19 & 3.26 & 17.55 & 48.71 \\
\normalsize \textbf{AceGPT-v2-8b-chat} & 55.25 & 73.33 & 53.14 & 62.50 & 58.39 & 39.82 & 41.06 & 47.16 & 93.33 & 24.59 & 49.39 & 77.57 & 22.47 & 44.97 & 66.30 & 4.80 & 23.52 & 49.33 \\
\normalsize \textbf{gemma-2-9b-it} & 50.72 & 49.44 & 49.53 & 61.35 & 61.79 & 35.79 & 40.23 & 48.01 & 81.66 & 23.09 & 46.98 & 75.42 & 11.73 & 39.00 & 60.42 & 2.68 & 24.28 & 48.26 \\
\normalsize \textbf{gemma-3-12b-it} & \underline{61.55} & \underline{77.00} & 49.49 & 64.96 & \textbf{63.53} & 38.03 & 41.27 & 48.86 & 92.61 & 22.90 & 45.97 & 73.46 & 5.24 & 32.82 & 54.34 & 2.77 & 26.16 & 50.47 \\
\normalsize \textbf{jais-family-13b-chat} & 44.85 & 66.33 & 52.99 & 64.85 & 57.91 & 36.91 & 33.26 & 38.64 & 52.52 & 10.41 & 31.98 & 64.15 & 8.64 & 30.10 & 57.00 & 0.84 & 11.35 & 44.71 \\
\normalsize \textbf{jais-adapted-13b-chat} & 50.03 & 65.33 & 47.53 & 61.30 & 56.72 & 37.14 & 35.45 & 41.76 & 52.91 & 15.53 & 41.48 & 70.86 & 15.96 & 38.81 & 63.52 & 1.00 & 13.33 & 46.08 \\
\normalsize \textbf{Qwen2.5-14B-Instruct} & 60.81 & 72.33 & 55.84 & 63.97 & 59.97 & 38.26 & 43.25 & 50.28 & 71.35 & 21.71 & 45.55 & 73.36 & 9.26 & 34.21 & 53.89 & 4.07 & 25.83 & 51.41 \\
\midrule[0.8pt]
\normalsize \textbf{Nile-Chat-4B} & 50.25 & 68.56 & 55.92 & 67.30 & 61.87 & 40.94 & 42.10 & 46.02 & 86.95 & 37.49 & 58.40 & 84.30 & 30.35 & 52.01 & 74.07 & 51.46 & 80.44 & 89.59 \\
\normalsize \textbf{Nile-Chat-2x4B-A6B} & 52.05 & 73.89 & \underline{59.69} & 68.67 & \underline{62.26} & \underline{41.61} & 44.07 & \underline{51.14} & \underline{94.58} & \underline{41.98} & \underline{61.59} & \underline{86.11} & \underline{33.40} & \underline{53.71} & \underline{76.78} & \underline{57.75} & \underline{83.89} & \underline{91.05} \\
\normalsize \textbf{Nile-Chat-3x4B-A6B} & 52.13 & 75.44 & 59.30 & \underline{69.27} & 57.91 & 41.16 & \underline{44.59} & 48.30 & 94.18 & \textbf{42.43} & \textbf{61.90} & \textbf{86.26} & \textbf{34.56} & \textbf{55.37} & \textbf{76.97} & \textbf{57.79} & \textbf{83.97} & \textbf{91.13} \\
\normalsize \textbf{Nile-Chat-12B} & \textbf{62.59} & \textbf{79.44} & \textbf{64.04} & \textbf{70.69} & \textbf{63.53} & \textbf{42.06} & \textbf{48.02} & \textbf{53.13} & \textbf{95.56} & 40.53 & 60.61 & 85.45 & 32.20 & 53.53 & 74.72 & 52.21 & 80.97 & 89.71 \\
\bottomrule
\end{tabular}}
\caption{Performance comparison of Nile-Chat and state-of-the-art models on the \textbf{Arabic-script}  benchmarks. The highest scores are indicated in \textbf{bold}, the second-highest are \underline{underlined}. Figure \ref{fig:overall-accuracy} shows the average score over all the benchmarks and measures for each model.}
\label{tab:arabic-results}
\end{table*}

\section{Evaluation Benchmarks}
To evaluate the performance of our models, we created eight benchmarks by translating widely used English LLM benchmarks into Egyptian Arabic using Claude, with four of them also rendered in the Latin script.
Additionally, we evaluated using held-out test sets from our translation and transliteration datasets (see Section~\ref{sec:instruction_datasets}), collectively referred to as \textbf{EgyptianBench}\footnote{\url{https://hf.co/datasets/MBZUAI-Paris/EgyptianBench}}.
All our custom benchmarks are integrated into a fork\footnote{\url{https://github.com/MBZUAI-Paris/lm-evaluation-harness-nile-chat}} of the LM-Evaluation-Harness repository \citep{eval-harness} to ensure reproducibility and foster future comparison.

\smallskip
\noindent\textbf{EgyptianMMLU}\footnote{\url{https://hf.co/datasets/MBZUAI-Paris/EgyptianMMLU}}. We combined two sources: \textit{ArabicMMLU-egy} \citep{mousi-etal-2025-aradice}, an Egyptian translated version of ArabicMMLU \citep{koto-etal-2024-arabicmmlu} using an in-house dialect translation system and subsequently validated by human annotators, and \textit{English MMLU} \cite{hendrycks2020measuring}, which we translated directly into Egyptian.

\smallskip
\noindent\textbf{Belebele-Arz} \citep{bandarkar2023belebele}. It is a multiple-choice machine reading comprehension benchmark across many languages. We adopted the provided Egyptian Arabic subset directly.

\smallskip
\noindent\textbf{EgyptianHellaSwag} \citep{zellers2019hellaswag}\footnote{\url{https://hf.co/datasets/MBZUAI-Paris/EgyptianHellaSwag}}. It presents complex scenarios where models must select the most plausible continuation of a given context from four options, challenging nuanced language understanding and contextual inference.

\smallskip
\noindent\textbf{EgyptianPIQA}~\citep{bisk2020piqa}\footnote{\url{https://hf.co/datasets/MBZUAI-Paris/EgyptianPIQA}}. The Physical Interaction Question Answering (PIQA) evaluates physical commonsense reasoning, presenting pairs of \emph{Goal} and \emph{Solution} options about everyday interactions with the physical world.

\smallskip
\noindent\textbf{EgyptianWinoGrande}~\citep{sakaguchi2021winogrande}\footnote{\url{https://hf.co/datasets/MBZUAI-Paris/EgyptianWinoGrande}}. It consists of fill-in-the-blank coreference problems where models must choose the correct noun phrase to resolve an ambiguous pronoun, a task demanding nuanced commonsense reasoning.

\smallskip
\noindent\textbf{EgyptianOpenBookQA} \citep{mihaylov2018can}\footnote{\url{https://hf.co/datasets/MBZUAI-Paris/EgyptianOpenBookQA}}.
This benchmark contains elementary-level science questions that require both explicit facts and broader commonsense knowledge; in translating it to Egyptian Arabic, we preserved scientific terminology to keep the questions accurate.

\smallskip
\noindent\textbf{EgyptianRACE} \citep{lai2017race}\footnote{\url{https://hf.co/datasets/MBZUAI-Paris/EgyptianRACE}}. ReAding ComprEhension (RACE) consists of English exam questions for middle and high school students, evaluating cognitive skills including reading comprehension, summarization, inference, and reasoning. In translating it to Egyptian Arabic, we preserved its narrative structure and question integrity.

\smallskip

\noindent\textbf{EgyptianAlpacaEval} \citep{dubois2024length}\footnote{\url{https://hf.co/datasets/MBZUAI-Paris/EgyptianAlpacaEval}}. AlpacaEval is designed to evaluate instruction-following capabilities via pairwise comparison. We adapted this framework to Egyptian Arabic by constructing a culturally grounded evaluation set in the Arabic script. In this setting, a judge model compares two responses generated by different models for the same prompt and selects the one that best aligns with Egyptian linguistic norms, cultural values, and pragmatic appropriateness.

\section{Results}

\noindent\textbf{Evaluation Measures}.
We used \textit{accuracy} as the evaluation metric across all multiple-choice QA benchmarks, except for EgyptianHellaSwag, we adopted \textit{normalized accuracy}. For translation and transliteration tasks, we used BLEU and chrF to evaluate surface-level correspondence, and BERTScore to assess the semantic similarity between the model outputs and the reference texts. Specifically, for BERTScore computation, we used multilingual BERT (mBERT) \citep{devlin-etal-2019-bert} for translations into Egyptian Arabic, AraBERT \citep{antoun-etal-2020-arabert} for translations into MSA, and BERT-base for translations into English. For the transliteration tasks in both directions (Arabic to Latin and Latin to Arabic), we used mBERT. 

The EgyptianAlpacaEval uses an LLM-as-a-Judge approach \citep{zheng2023judging}, where Claude is tasked with selecting the more culturally appropriate response between two candidates. We used \texttt{AceGPT-v1.5-13B-Chat} \citep{zhu2024second} as the reference model. We generated the candidate outputs using the default sampling-based decoding for each model. We applied the chat template for all benchmarks, except for EgyptianWinoGrande.

\begin{table}[t]
\centering
\renewcommand{\arraystretch}{1.13}
\setlength{\tabcolsep}{4pt}
\resizebox{\columnwidth}{!}{
\begin{tabular}{p{4cm}ccccc}
\toprule
\textbf{\small Model} &
\shortstack{\textbf{\small Egyptian} \\ \textbf{\small HellaSwag}} &
\shortstack{\textbf{\small Egyptian} \\ \textbf{\small PIQA}} &
\shortstack{\textbf{\small Egyptian} \\ \textbf{\small WinoGrande}} &
\shortstack{\textbf{\small Egyptian} \\ \textbf{\small RACE-H}} &
\shortstack{\textbf{\small Egyptian} \\ \textbf{\small RACE-M}} \\
\midrule
\small \textbf{gemma-3-4b-it}                   & 30.90 & 52.76 & 48.57 & 25.47 & 26.94 \\
\small \textbf{jais-family-6p7b-chat}           & 30.27 & 53.25 & 52.14 & 24.18 & 28.06 \\
\small \textbf{jais-adapted-7b-chat}            & 30.81 & 51.67 & 50.40 & 24.38 & 28.06 \\
\small \textbf{Qwen2.5-7B-Instruct}             & 30.51 & 51.88 & 50.95 & 24.88 & 26.11 \\
\small \textbf{ALLaM-7B-Instruct-preview}       & 32.17 & 53.09 & 50.63 & 25.07 & 31.94 \\
\small \textbf{c4ai-command-r7b-arabic-02-2025} & 30.88 & 52.32 & 51.43 & 25.07 & 27.22 \\
\small \textbf{Llama-3.1-8B-Instruct}           & 31.77 & 53.30 & 50.24 & 24.48 & 28.33 \\
\small \textbf{AceGPT-v2-8b-chat}               & 33.16 & 53.80 & 50.24 & 26.07 & 30.56 \\
\small \textbf{gemma-2-9b-it}                   & 33.75 & 53.69 & 50.79 & 26.66 & 28.61 \\
\small \textbf{gemma-3-12b-it}                  & 37.52 & 53.14 & 51.19 & 31.02 & 35.28 \\
\small \textbf{jais-family-13b-chat}            & 30.46 & 53.09 & 48.18 & 25.28 & 27.78 \\
\small \textbf{jais-adapted-13b-chat}           & 31.14 & 52.87 & 50.79 & 23.98 & 26.11 \\
\small \textbf{Qwen2.5-14B-Instruct}            & 33.49 & 52.87 & 53.41 & 27.35 & 30.28 \\
\midrule[0.8pt]     
\small \textbf{Nile-Chat-4B}                    & 50.55 & 65.32 & \underline{60.62} & 37.36 & 43.06 \\
\small \textbf{Nile-Chat-2x4B-A6B}              & \textbf{55.49} & \textbf{68.00} & \textbf{61.33} & 40.24 & \underline{45.56} \\
\small \textbf{Nile-Chat-3x4B-A6B}              & \underline{55.00} & \underline{66.68} & 56.42 & \underline{40.44} & 42.78 \\
\small \textbf{Nile-Chat-12B}                   & 53.71 & 65.10 & 59.98 & \textbf{41.72} & \textbf{48.89} \\
\bottomrule
\end{tabular}}
\caption{Performance comparison of Nile-Chat and state-of-the-art models on the \textbf{Latin-script} benchmarks.}
\label{tab:latin-results}
\end{table}

\smallskip

\noindent\textbf{Result Analysis.}
The evaluation results in Tables \ref{tab:arabic-results} and \ref{tab:latin-results} demonstrate the exceptional performance of the Nile-Chat models across all Egyptian benchmarks in both the Arabic and the Latin scripts. 

Compared to models with 7B parameters or fewer, \textbf{\texttt{Nile-Chat-4B}} demonstrates consistently superior performance across multiple \textit{Arabic-script} benchmarks, achieving relative gains of 1.2\% on EgyptianPIQA, 0.9\% on EgyptianOpenBookQA, 0.21\% on EgyptianRACE-High, and 1.6\% on EgyptianAlpacaEval over the strongest competitor for each task. It also ranks first in translation and transliteration tasks across all evaluation metrics. On the \textit{Latin-script} benchmarks, \texttt{4B} outperforms all models in the same size category, by sizable margins: +18.38\% on EgyptianHellaSwag, +12.97\% on EgyptianPIQA, +8.48\% on EgyptianWinoGrande, +11.91\% on EgyptianRACE-High, and +11.12\% on EgyptianRACE Medium, relative to the next-best model. \textit{This indicates that existing LLMs underrepresent or overlook the Latin script.}

\textbf{\texttt{Nile-Chat-12B}}, on the other hand, pushes the state-of-the-art even further. Across the \textit{Arabic-script} benchmarks, it achieves the highest score on every task, with the largest absolute improvements of +4.35\% on EgyptianHellaSwag and +3.43\% on EgyptianRACE-High over the next-best model. It also performs exceptionally well on the \textit{Latin-script} and generation benchmarks, leading on EgyptianRACE-High (+1.28\%) and EgyptianRACE-Medium (+3.33\%), and ranking consistently within 1–3\% of the top-performing models on the remaining Latin tasks, translation, and transliteration metrics. In all such cases, the models that marginally outperformed it belong to the MoE-based Nile-Chat family.

\textbf{\texttt{Nile-Chat-3x4B-A6B}} and \textbf{\texttt{2x4B-A6B}} strike a balance between the \texttt{4B} and \texttt{12B} dense models on discriminative \textit{Arabic-script} tasks, yet excel whenever extensive generation or \textit{Latin-script} processing is required. On EgyptianHellaSwag, they score 59.69\% and 59.30\%, respectively, which  ranks them between the dense 4B (55.92\%) and 12B (64.04\%) models. A similar pattern holds for EgyptianPIQA.  In \textit{Latin-script}, \texttt{2x4B-A6B} leads three of five benchmarks, widening the gap with the \texttt{4B} dense model by 4.94\% on EgyptianHellaSwag and 2.68\% on EgyptianPIQA, while keeping within approximately 1–3\% of the \texttt{12B} model on the Latin RACE tasks.   For generation tasks, \texttt{3x4B-A6B} achieves the highest scores across \emph{all} translation and transliteration tasks and metrics.

\begin{figure}[t]
    \centering
    \includegraphics[width=1\linewidth]{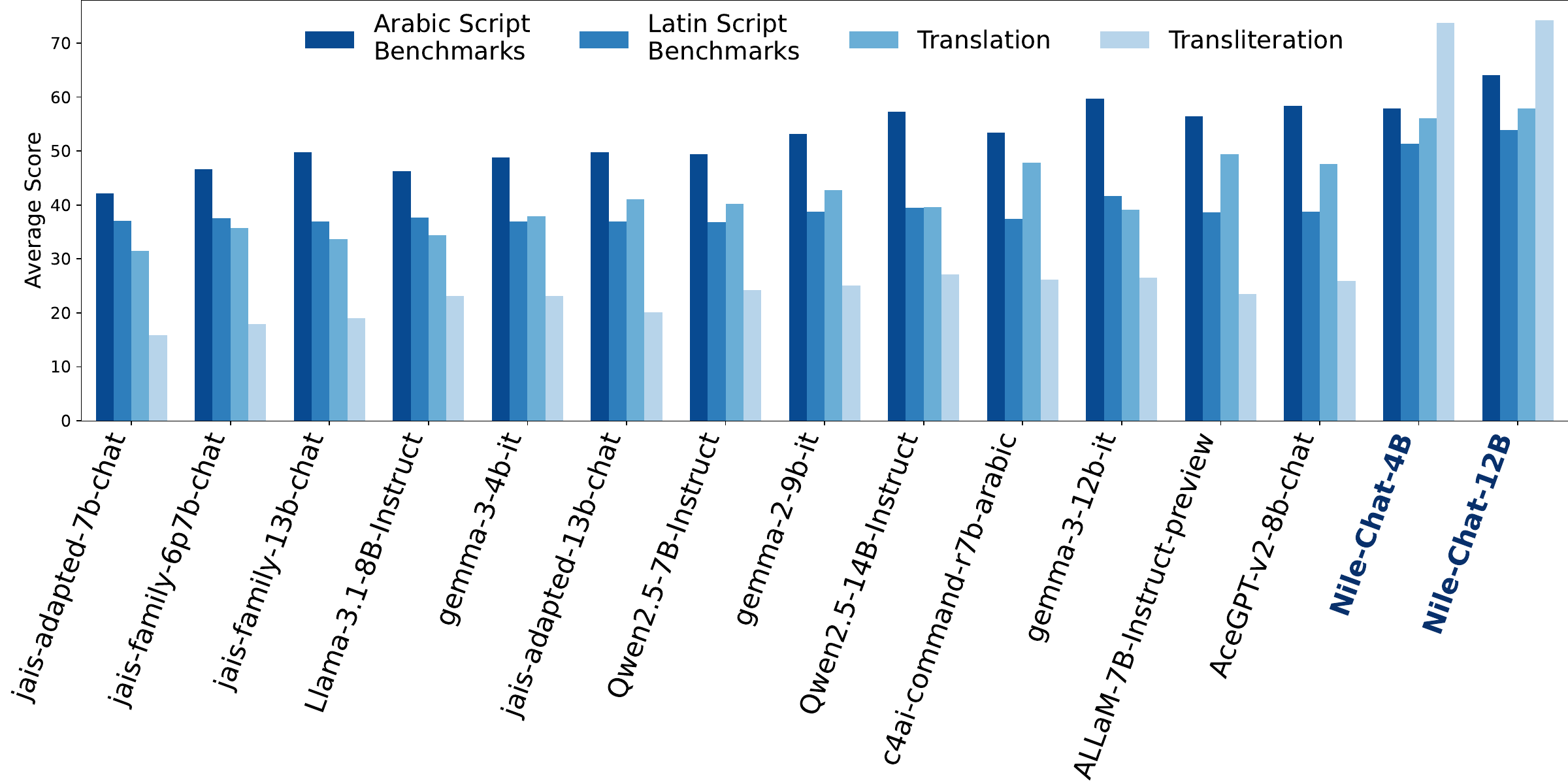}
    \caption{Average model scores over the benchmarks.}
    \label{fig:overall-accuracy}
\end{figure}

\section{Conclusion}
We introduced Nile-Chat, a family of language models specifically designed for the Egyptian Arabic dialect, uniquely capable of understanding and generating texts in both Arabic and Latin scripts. Our novel Branch-Train-MiX (BTX) based MoE model effectively integrates script-specialized experts, demonstrating superior performance across various benchmarks compared to leading multilingual and Arabic-specific models. Nile-Chat significantly enhances LLM capabilities in dual-script settings, achieving sizable improvement over current state-of-the-art models on Latin-script tasks. By releasing all our resources, datasets, and evaluation suites publicly, we aim to encourage further research and development in dual-script language modeling, addressing critical gaps for widely spoken yet underrepresented languages.

\section*{Limitations}
Despite the promising results, our work has some limitations. First, the model occasionally generates hallucinations. Second, the dataset may contain inherent biases that could affect the model’s fairness and representation. Additionally, we relied heavily on Claude for translating English instructions into Egyptian Arabic. However, because Claude is primarily trained on English and reflects Western cultural values, it may not fully capture the unique nuances of Egyptian Arabic. We intend to address these limitations in future work.

\bibliography{googlescholar.bib,aclanthology.bib}

\begin{thebibliography}{60}
\providecommand{\natexlab}[1]{#1}

\bibitem[{Al-Sabbagh(2024)}]{al2024arzen}
Rania Al-Sabbagh. 2024.
\newblock Arzen-multigenre: An aligned parallel dataset of egyptian arabic song lyrics, novels, and subtitles, with english translations.
\newblock \emph{Data in Brief}, 54:110271.

\bibitem[{Alghamdi(2018)}]{alghamdi2018arabizi}
Hamdah~Abdullah Alghamdi. 2018.
\newblock \emph{Arabizi: An exploration of the use of the contemporary youth netspeak on Social Networking Sites in Saudi Arabia}.
\newblock Ph.D. thesis, University of Canberra.

\bibitem[{Antoun et~al.(2020)Antoun, Baly, and Hajj}]{antoun-etal-2020-arabert}
Wissam Antoun, Fady Baly, and Hazem Hajj. 2020.
\newblock \href {https://aclanthology.org/2020.osact-1.2/} {{A}ra{BERT}: Transformer-based model for {A}rabic language understanding}.
\newblock In \emph{Proceedings of the 4th Workshop on Open-Source Arabic Corpora and Processing Tools, with a Shared Task on Offensive Language Detection}, pages 9--15, Marseille, France. European Language Resource Association.

\bibitem[{Bandarkar et~al.(2023)Bandarkar, Liang, Muller, Artetxe, Shukla, Husa, Goyal, Krishnan, Zettlemoyer, and Khabsa}]{bandarkar2023belebele}
Lucas Bandarkar, Davis Liang, Benjamin Muller, Mikel Artetxe, Satya~Narayan Shukla, Donald Husa, Naman Goyal, Abhinandan Krishnan, Luke Zettlemoyer, and Madian Khabsa. 2023.
\newblock The belebele benchmark: a parallel reading comprehension dataset in 122 language variants.
\newblock \emph{arXiv preprint arXiv:2308.16884}.

\bibitem[{Bari et~al.(2024)Bari, Alnumay, Alzahrani, Alotaibi, Alyahya, AlRashed, Mirza, Alsubaie, Alahmed, Alabduljabbar et~al.}]{bari2024allam}
M~Saiful Bari, Yazeed Alnumay, Norah~A Alzahrani, Nouf~M Alotaibi, Hisham~A Alyahya, Sultan AlRashed, Faisal~A Mirza, Shaykhah~Z Alsubaie, Hassan~A Alahmed, Ghadah Alabduljabbar, and 1 others. 2024.
\newblock Allam: Large language models for arabic and english.
\newblock \emph{arXiv preprint arXiv:2407.15390}.

\bibitem[{Bisk et~al.(2020)Bisk, Zellers, Gao, Choi et~al.}]{bisk2020piqa}
Yonatan Bisk, Rowan Zellers, Jianfeng Gao, Yejin Choi, and 1 others. 2020.
\newblock Piqa: Reasoning about physical commonsense in natural language.
\newblock In \emph{Proceedings of the AAAI conference on artificial intelligence}, volume~34, pages 7432--7439.

\bibitem[{Brahman et~al.(2024)Brahman, Kumar, Balachandran, Dasigi, Pyatkin, Ravichander, Wiegreffe, Dziri, Chandu, Hessel et~al.}]{brahman2024art}
Faeze Brahman, Sachin Kumar, Vidhisha Balachandran, Pradeep Dasigi, Valentina Pyatkin, Abhilasha Ravichander, Sarah Wiegreffe, Nouha Dziri, Khyathi Chandu, Jack Hessel, and 1 others. 2024.
\newblock The art of saying no: Contextual noncompliance in language models.
\newblock \emph{Advances in Neural Information Processing Systems}, 37:49706--49748.

\bibitem[{Chen et~al.(2024)Chen, Li, Dong, Zhang, He, Wang, Zhao, and Lin}]{chen2024sharegpt4v}
Lin Chen, Jinsong Li, Xiaoyi Dong, Pan Zhang, Conghui He, Jiaqi Wang, Feng Zhao, and Dahua Lin. 2024.
\newblock Sharegpt4v: Improving large multi-modal models with better captions.
\newblock In \emph{European Conference on Computer Vision}, pages 370--387. Springer.

\bibitem[{Choudhury et~al.(2025)Choudhury, Chauhan, Das, Sahnan, Han, Li, Singh, Jadhav, Agarwal, Choudhary et~al.}]{choudhury2025llama}
Monojit Choudhury, Shivam Chauhan, Rocktim~Jyoti Das, Dhruv Sahnan, Xudong Han, Haonan Li, Aaryamonvikram Singh, Alok~Anil Jadhav, Utkarsh Agarwal, Mukund Choudhary, and 1 others. 2025.
\newblock Llama-3-nanda-10b-chat: An open generative large language model for hindi.
\newblock \emph{arXiv preprint arXiv:2504.06011}.

\bibitem[{Costa-Juss{\`a} et~al.(2022)Costa-Juss{\`a}, Cross, {\c{C}}elebi, Elbayad, Heafield, Heffernan, Kalbassi, Lam, Licht, Maillard et~al.}]{costa2022no}
Marta~R Costa-Juss{\`a}, James Cross, Onur {\c{C}}elebi, Maha Elbayad, Kenneth Heafield, Kevin Heffernan, Elahe Kalbassi, Janice Lam, Daniel Licht, Jean Maillard, and 1 others. 2022.
\newblock No language left behind: Scaling human-centered machine translation.
\newblock \emph{arXiv preprint arXiv:2207.04672}.

\bibitem[{Darwish(2013)}]{darwish2013arabizi}
Kareem Darwish. 2013.
\newblock Arabizi detection and conversion to arabic.
\newblock \emph{arXiv preprint arXiv:1306.6755}.

\bibitem[{Deng et~al.(2024)Deng, Zhao, Hessel, Ren, Cardie, and Choi}]{deng2024wildvis}
Yuntian Deng, Wenting Zhao, Jack Hessel, Xiang Ren, Claire Cardie, and Yejin Choi. 2024.
\newblock Wildvis: Open source visualizer for million-scale chat logs in the wild.
\newblock \emph{arXiv preprint arXiv:2409.03753}.

\bibitem[{Devlin et~al.(2019)Devlin, Chang, Lee, and Toutanova}]{devlin-etal-2019-bert}
Jacob Devlin, Ming-Wei Chang, Kenton Lee, and Kristina Toutanova. 2019.
\newblock \href {https://doi.org/10.18653/v1/N19-1423} {{BERT}: Pre-training of deep bidirectional transformers for language understanding}.
\newblock In \emph{Proceedings of the 2019 Conference of the North {A}merican Chapter of the Association for Computational Linguistics: Human Language Technologies, Volume 1 (Long and Short Papers)}, pages 4171--4186, Minneapolis, Minnesota. Association for Computational Linguistics.

\bibitem[{Ding et~al.(2023)Ding, Chen, Xu, Qin, Zheng, Hu, Liu, Sun, and Zhou}]{ding2023enhancing}
Ning Ding, Yulin Chen, Bokai Xu, Yujia Qin, Zhi Zheng, Shengding Hu, Zhiyuan Liu, Maosong Sun, and Bowen Zhou. 2023.
\newblock Enhancing chat language models by scaling high-quality instructional conversations.
\newblock \emph{arXiv preprint arXiv:2305.14233}.

\bibitem[{Dubois et~al.(2024)Dubois, Galambosi, Liang, and Hashimoto}]{dubois2024length}
Yann Dubois, Bal{\'a}zs Galambosi, Percy Liang, and Tatsunori~B Hashimoto. 2024.
\newblock Length-controlled alpacaeval: A simple way to debias automatic evaluators.
\newblock \emph{arXiv preprint arXiv:2404.04475}.

\bibitem[{El-Haj(2020)}]{el-haj-2020-habibi}
Mahmoud El-Haj. 2020.
\newblock \href {https://aclanthology.org/2020.lrec-1.165/} {Habibi - a multi dialect multi national {A}rabic song lyrics corpus}.
\newblock In \emph{Proceedings of the Twelfth Language Resources and Evaluation Conference}, pages 1318--1326, Marseille, France. European Language Resources Association.

\bibitem[{Faheem et~al.(2024)Faheem, Wassif, Bayomi, and Abdou}]{faheem2024improving}
Mohamed~Atta Faheem, Khaled~Tawfik Wassif, Hanaa Bayomi, and Sherif~Mahdy Abdou. 2024.
\newblock Improving neural machine translation for low resource languages through non-parallel corpora: a case study of egyptian dialect to modern standard arabic translation.
\newblock \emph{Scientific Reports}, 14(1):2265.

\bibitem[{Gao et~al.(2024)Gao, Tow, Abbasi, Biderman, Black, DiPofi, Foster, Golding, Hsu, Le~Noac'h, Li, McDonell, Muennighoff, Ociepa, Phang, Reynolds, Schoelkopf, Skowron, Sutawika, Tang, Thite, Wang, Wang, and Zou}]{eval-harness}
Leo Gao, Jonathan Tow, Baber Abbasi, Stella Biderman, Sid Black, Anthony DiPofi, Charles Foster, Laurence Golding, Jeffrey Hsu, Alain Le~Noac'h, Haonan Li, Kyle McDonell, Niklas Muennighoff, Chris Ociepa, Jason Phang, Laria Reynolds, Hailey Schoelkopf, Aviya Skowron, Lintang Sutawika, and 5 others. 2024.
\newblock \href {https://doi.org/10.5281/zenodo.12608602} {A framework for few-shot language model evaluation}.

\bibitem[{Han et~al.(2024)Han, Rao, Ettinger, Jiang, Lin, Lambert, Choi, and Dziri}]{han2024wildguard}
Seungju Han, Kavel Rao, Allyson Ettinger, Liwei Jiang, Bill~Yuchen Lin, Nathan Lambert, Yejin Choi, and Nouha Dziri. 2024.
\newblock Wildguard: Open one-stop moderation tools for safety risks, jailbreaks, and refusals of llms.
\newblock \emph{arXiv preprint arXiv:2406.18495}.

\bibitem[{Hendrycks et~al.(2020)Hendrycks, Burns, Basart, Zou, Mazeika, Song, and Steinhardt}]{hendrycks2020measuring}
Dan Hendrycks, Collin Burns, Steven Basart, Andy Zou, Mantas Mazeika, Dawn Song, and Jacob Steinhardt. 2020.
\newblock Measuring massive multitask language understanding.
\newblock \emph{arXiv preprint arXiv:2009.03300}.

\bibitem[{Huang et~al.(2024)Huang, Yu, Zhu, Sun, Cheng, Dingjie, Chen, Alharthi, An, He, Liu, Chen, Li, Wang, Zhang, Sun, Wan, Li, and Xu}]{huang-etal-2024-acegpt}
Huang Huang, Fei Yu, Jianqing Zhu, Xuening Sun, Hao Cheng, Song Dingjie, Zhihong Chen, Mosen Alharthi, Bang An, Juncai He, Ziche Liu, Junying Chen, Jianquan Li, Benyou Wang, Lian Zhang, Ruoyu Sun, Xiang Wan, Haizhou Li, and Jinchao Xu. 2024.
\newblock \href {https://doi.org/10.18653/v1/2024.naacl-long.450} {{A}ce{GPT}, localizing large language models in {A}rabic}.
\newblock In \emph{Proceedings of the 2024 Conference of the North American Chapter of the Association for Computational Linguistics: Human Language Technologies (Volume 1: Long Papers)}, pages 8139--8163, Mexico City, Mexico. Association for Computational Linguistics.

\bibitem[{Ivison et~al.(2023)Ivison, Wang, Pyatkin, Lambert, Peters, Dasigi, Jang, Wadden, Smith, Beltagy et~al.}]{ivison2023camels}
Hamish Ivison, Yizhong Wang, Valentina Pyatkin, Nathan Lambert, Matthew Peters, Pradeep Dasigi, Joel Jang, David Wadden, Noah~A Smith, Iz~Beltagy, and 1 others. 2023.
\newblock Camels in a changing climate: Enhancing lm adaptation with tulu 2.
\newblock \emph{arXiv preprint arXiv:2311.10702}.

\bibitem[{Jiang et~al.(2024{\natexlab{a}})Jiang, Sablayrolles, Roux, Mensch, Savary, Bamford, Chaplot, Casas, Hanna, Bressand et~al.}]{jiang2024mixtral}
Albert~Q Jiang, Alexandre Sablayrolles, Antoine Roux, Arthur Mensch, Blanche Savary, Chris Bamford, Devendra~Singh Chaplot, Diego de~las Casas, Emma~Bou Hanna, Florian Bressand, and 1 others. 2024{\natexlab{a}}.
\newblock Mixtral of experts.
\newblock \emph{arXiv preprint arXiv:2401.04088}.

\bibitem[{Jiang et~al.(2024{\natexlab{b}})Jiang, Rao, Han, Ettinger, Brahman, Kumar, Mireshghallah, Lu, Sap, Choi et~al.}]{jiang2024wildteaming}
Liwei Jiang, Kavel Rao, Seungju Han, Allyson Ettinger, Faeze Brahman, Sachin Kumar, Niloofar Mireshghallah, Ximing Lu, Maarten Sap, Yejin Choi, and 1 others. 2024{\natexlab{b}}.
\newblock Wildteaming at scale: From in-the-wild jailbreaks to (adversarially) safer language models.
\newblock \emph{Advances in Neural Information Processing Systems}, 37:47094--47165.

\bibitem[{Kargaran et~al.(2023)Kargaran, Imani, Yvon, and Schuetze}]{kargaran-etal-2023-glotlid}
Amir~Hossein Kargaran, Ayyoob Imani, Fran{\c{c}}ois Yvon, and Hinrich Schuetze. 2023.
\newblock \href {https://doi.org/10.18653/v1/2023.findings-emnlp.410} {{G}lot{LID}: Language identification for low-resource languages}.
\newblock In \emph{Findings of the Association for Computational Linguistics: EMNLP 2023}, pages 6155--6218, Singapore. Association for Computational Linguistics.

\bibitem[{K{\"o}pf et~al.(2023)K{\"o}pf, Kilcher, Von~R{\"u}tte, Anagnostidis, Tam, Stevens, Barhoum, Nguyen, Stanley, Nagyfi et~al.}]{kopf2023openassistant}
Andreas K{\"o}pf, Yannic Kilcher, Dimitri Von~R{\"u}tte, Sotiris Anagnostidis, Zhi~Rui Tam, Keith Stevens, Abdullah Barhoum, Duc Nguyen, Oliver Stanley, Rich{\'a}rd Nagyfi, and 1 others. 2023.
\newblock Openassistant conversations-democratizing large language model alignment.
\newblock \emph{Advances in Neural Information Processing Systems}, 36:47669--47681.

\bibitem[{Koto et~al.(2025)Koto, Joshi, Mukhituly, Wang, Xie, Pal, Orel, Mullah, Turmakhan, Goloburda et~al.}]{koto2025llama}
Fajri Koto, Rituraj Joshi, Nurdaulet Mukhituly, Yuxia Wang, Zhuohan Xie, Rahul Pal, Daniil Orel, Parvez Mullah, Diana Turmakhan, Maiya Goloburda, and 1 others. 2025.
\newblock Llama-3.1-sherkala-8b-chat: An open large language model for kazakh.
\newblock \emph{arXiv preprint arXiv:2503.01493}.

\bibitem[{Koto et~al.(2024)Koto, Li, Shatnawi, Doughman, Sadallah, Alraeesi, Almubarak, Alyafeai, Sengupta, Shehata, Habash, Nakov, and Baldwin}]{koto-etal-2024-arabicmmlu}
Fajri Koto, Haonan Li, Sara Shatnawi, Jad Doughman, Abdelrahman Sadallah, Aisha Alraeesi, Khalid Almubarak, Zaid Alyafeai, Neha Sengupta, Shady Shehata, Nizar Habash, Preslav Nakov, and Timothy Baldwin. 2024.
\newblock \href {https://doi.org/10.18653/v1/2024.findings-acl.334} {{A}rabic{MMLU}: Assessing massive multitask language understanding in {A}rabic}.
\newblock In \emph{Findings of the Association for Computational Linguistics: ACL 2024}, pages 5622--5640, Bangkok, Thailand. Association for Computational Linguistics.

\bibitem[{Lai et~al.(2017)Lai, Xie, Liu, Yang, and Hovy}]{lai2017race}
Guokun Lai, Qizhe Xie, Hanxiao Liu, Yiming Yang, and Eduard Hovy. 2017.
\newblock Race: Large-scale reading comprehension dataset from examinations.
\newblock \emph{arXiv preprint arXiv:1704.04683}.

\bibitem[{Lambert et~al.(2024)Lambert, Morrison, Pyatkin, Huang, Ivison, Brahman, Miranda, Liu, Dziri, Lyu et~al.}]{lambert2024t}
Nathan Lambert, Jacob Morrison, Valentina Pyatkin, Shengyi Huang, Hamish Ivison, Faeze Brahman, Lester James~V Miranda, Alisa Liu, Nouha Dziri, Shane Lyu, and 1 others. 2024.
\newblock T$\backslash$" ulu 3: Pushing frontiers in open language model post-training.
\newblock \emph{arXiv preprint arXiv:2411.15124}.

\bibitem[{Li et~al.(2024{\natexlab{a}})Li, Lin, Duan, Liang, and Shroff}]{li2024theory}
Hongbo Li, Sen Lin, Lingjie Duan, Yingbin Liang, and Ness~B Shroff. 2024{\natexlab{a}}.
\newblock Theory on mixture-of-experts in continual learning.
\newblock \emph{arXiv preprint arXiv:2406.16437}.

\bibitem[{Li et~al.(2024{\natexlab{b}})Li, Beeching, Tunstall, Lipkin, Soletskyi, Huang, Rasul, Yu, Jiang, Shen et~al.}]{li2024numinamath}
Jia Li, Edward Beeching, Lewis Tunstall, Ben Lipkin, Roman Soletskyi, Shengyi Huang, Kashif Rasul, Longhui Yu, Albert~Q Jiang, Ziju Shen, and 1 others. 2024{\natexlab{b}}.
\newblock Numinamath: The largest public dataset in ai4maths with 860k pairs of competition math problems and solutions.
\newblock \emph{Hugging Face repository}, 13:9.

\bibitem[{Li et~al.(2023)Li, He, Yashar, Cui, Ge, Zhang, Fainman, Zhang, and Chaudhuri}]{li2023table}
Peng Li, Yeye He, Dror Yashar, Weiwei Cui, Song Ge, Haidong Zhang, Danielle~Rifinski Fainman, Dongmei Zhang, and Surajit Chaudhuri. 2023.
\newblock Table-gpt: Table-tuned gpt for diverse table tasks.
\newblock \emph{arXiv preprint arXiv:2310.09263}.

\bibitem[{Lo et~al.(2024)Lo, Huang, Qiu, Wang, and Fu}]{lo2024closer}
Ka~Man Lo, Zeyu Huang, Zihan Qiu, Zili Wang, and Jie Fu. 2024.
\newblock A closer look into mixture-of-experts in large language models.
\newblock \emph{arXiv preprint arXiv:2406.18219}.

\bibitem[{Longpre et~al.(2023)Longpre, Hou, Vu, Webson, Chung, Tay, Zhou, Le, Zoph, Wei et~al.}]{longpre2023flan}
Shayne Longpre, Le~Hou, Tu~Vu, Albert Webson, Hyung~Won Chung, Yi~Tay, Denny Zhou, Quoc~V Le, Barret Zoph, Jason Wei, and 1 others. 2023.
\newblock The flan collection: Designing data and methods for effective instruction tuning.
\newblock In \emph{International Conference on Machine Learning}, pages 22631--22648. PMLR.

\bibitem[{Luo et~al.(2023)Luo, Xu, Zhao, Sun, Geng, Hu, Tao, Ma, Lin, and Jiang}]{luo2023wizardcoder}
Ziyang Luo, Can Xu, Pu~Zhao, Qingfeng Sun, Xiubo Geng, Wenxiang Hu, Chongyang Tao, Jing Ma, Qingwei Lin, and Daxin Jiang. 2023.
\newblock Wizardcoder: Empowering code large language models with evol-instruct.
\newblock \emph{arXiv preprint arXiv:2306.08568}.

\bibitem[{Mihaylov et~al.(2018)Mihaylov, Clark, Khot, and Sabharwal}]{mihaylov2018can}
Todor Mihaylov, Peter Clark, Tushar Khot, and Ashish Sabharwal. 2018.
\newblock Can a suit of armor conduct electricity? a new dataset for open book question answering.
\newblock \emph{arXiv preprint arXiv:1809.02789}.

\bibitem[{Mohamed et~al.(2025)Mohamed, Zhang, Vazirgiannis, and Shang}]{mohamed2025lost}
Amr Mohamed, Yang Zhang, Michalis Vazirgiannis, and Guokan Shang. 2025.
\newblock Lost in the mix: Evaluating llm understanding of code-switched text.
\newblock \emph{arXiv preprint arXiv:2506.14012}.

\bibitem[{Mousi et~al.(2025)Mousi, Durrani, Ahmad, Hasan, Hasanain, Kabbani, Dalvi, Chowdhury, and Alam}]{mousi-etal-2025-aradice}
Basel Mousi, Nadir Durrani, Fatema Ahmad, Md.~Arid Hasan, Maram Hasanain, Tameem Kabbani, Fahim Dalvi, Shammur~Absar Chowdhury, and Firoj Alam. 2025.
\newblock \href {https://aclanthology.org/2025.coling-main.283/} {{A}ra{D}i{CE}: Benchmarks for dialectal and cultural capabilities in {LLM}s}.
\newblock In \emph{Proceedings of the 31st International Conference on Computational Linguistics}, pages 4186--4218, Abu Dhabi, UAE. Association for Computational Linguistics.

\bibitem[{Mukherjee et~al.(2023)Mukherjee, Mitra, Jawahar, Agarwal, Palangi, and Awadallah}]{mukherjee2023orca}
Subhabrata Mukherjee, Arindam Mitra, Ganesh Jawahar, Sahaj Agarwal, Hamid Palangi, and Ahmed Awadallah. 2023.
\newblock Orca: Progressive learning from complex explanation traces of gpt-4.
\newblock \emph{arXiv preprint arXiv:2306.02707}.

\bibitem[{Penedo et~al.(2025)Penedo, Kydl{\'\i}{\v{c}}ek, Sabol{\v{c}}ec, Messmer, Foroutan, Kargaran, Raffel, Jaggi, Von~Werra, and Wolf}]{penedo2025fineweb2}
Guilherme Penedo, Hynek Kydl{\'\i}{\v{c}}ek, Vinko Sabol{\v{c}}ec, Bettina Messmer, Negar Foroutan, Amir~Hossein Kargaran, Colin Raffel, Martin Jaggi, Leandro Von~Werra, and Thomas Wolf. 2025.
\newblock Fineweb2: One pipeline to scale them all--adapting pre-training data processing to every language.
\newblock \emph{arXiv preprint arXiv:2506.20920}.

\bibitem[{Peng et~al.(2023)Peng, Li, He, Galley, and Gao}]{peng2023instruction}
Baolin Peng, Chunyuan Li, Pengcheng He, Michel Galley, and Jianfeng Gao. 2023.
\newblock Instruction tuning with gpt-4.
\newblock \emph{arXiv preprint arXiv:2304.03277}.

\bibitem[{Qarah(2024)}]{qarah2024egybert}
Faisal Qarah. 2024.
\newblock Egybert: A large language model pretrained on egyptian dialect corpora.
\newblock \emph{arXiv preprint arXiv:2408.03524}.

\bibitem[{Rafailov et~al.(2023)Rafailov, Sharma, Mitchell, Manning, Ermon, and Finn}]{rafailov2023direct}
Rafael Rafailov, Archit Sharma, Eric Mitchell, Christopher~D Manning, Stefano Ermon, and Chelsea Finn. 2023.
\newblock Direct preference optimization: Your language model is secretly a reward model.
\newblock \emph{Advances in Neural Information Processing Systems}, 36:53728--53741.

\bibitem[{Robinson et~al.(2024)Robinson, Abdelmoneim, Marchisio, and Ruder}]{robinson2024qasida}
Nathaniel~R Robinson, Shahd Abdelmoneim, Kelly Marchisio, and Sebastian Ruder. 2024.
\newblock Al-qasida: Analyzing llm quality and accuracy systematically in dialectal arabic.
\newblock \emph{arXiv preprint arXiv:2412.04193}.

\bibitem[{Sakaguchi et~al.(2021)Sakaguchi, Bras, Bhagavatula, and Choi}]{sakaguchi2021winogrande}
Keisuke Sakaguchi, Ronan~Le Bras, Chandra Bhagavatula, and Yejin Choi. 2021.
\newblock Winogrande: An adversarial winograd schema challenge at scale.
\newblock \emph{Communications of the ACM}, 64(9):99--106.

\bibitem[{Sengupta et~al.(2023)Sengupta, Sahu, Jia, Katipomu, Li, Koto, Marshall, Gosal, Liu, Chen et~al.}]{sengupta2023jais}
Neha Sengupta, Sunil~Kumar Sahu, Bokang Jia, Satheesh Katipomu, Haonan Li, Fajri Koto, William Marshall, Gurpreet Gosal, Cynthia Liu, Zhiming Chen, and 1 others. 2023.
\newblock Jais and jais-chat: Arabic-centric foundation and instruction-tuned open generative large language models.
\newblock \emph{arXiv preprint arXiv:2308.16149}.

\bibitem[{Shang et~al.(2025)Shang, Abdine, Khoubrane, Mohamed, Abbahaddou, Ennadir, Momayiz, Ren, Moulines, Nakov, Vazirgiannis, and Xing}]{shang-etal-2025-atlas}
Guokan Shang, Hadi Abdine, Yousef Khoubrane, Amr Mohamed, Yassine Abbahaddou, Sofiane Ennadir, Imane Momayiz, Xuguang Ren, Eric Moulines, Preslav Nakov, Michalis Vazirgiannis, and Eric Xing. 2025.
\newblock \href {https://aclanthology.org/2025.loreslm-1.2/} {Atlas-chat: Adapting large language models for low-resource {M}oroccan {A}rabic dialect}.
\newblock In \emph{Proceedings of the First Workshop on Language Models for Low-Resource Languages}, pages 9--30, Abu Dhabi, United Arab Emirates. Association for Computational Linguistics.

\bibitem[{Singh et~al.(2024)Singh, Vargus, D{'}souza, Karlsson, Mahendiran, Ko, Shandilya, Patel, Mataciunas, O{'}Mahony, Zhang, Hettiarachchi, Wilson, Machado, Moura, Krzemi{\'n}ski, Fadaei, Ergun, Okoh, Alaagib, Mudannayake, Alyafeai, Chien, Ruder, Guthikonda, Alghamdi, Gehrmann, Muennighoff, Bartolo, Kreutzer, {\"U}st{\"u}n, Fadaee, and Hooker}]{singh-etal-2024-aya}
Shivalika Singh, Freddie Vargus, Daniel D{'}souza, B{\"o}rje~F. Karlsson, Abinaya Mahendiran, Wei-Yin Ko, Herumb Shandilya, Jay Patel, Deividas Mataciunas, Laura O{'}Mahony, Mike Zhang, Ramith Hettiarachchi, Joseph Wilson, Marina Machado, Luisa Moura, Dominik Krzemi{\'n}ski, Hakimeh Fadaei, Irem Ergun, Ifeoma Okoh, and 14 others. 2024.
\newblock \href {https://doi.org/10.18653/v1/2024.acl-long.620} {Aya dataset: An open-access collection for multilingual instruction tuning}.
\newblock In \emph{Proceedings of the 62nd Annual Meeting of the Association for Computational Linguistics (Volume 1: Long Papers)}, pages 11521--11567, Bangkok, Thailand. Association for Computational Linguistics.

\bibitem[{Sukhbaatar et~al.(2024)Sukhbaatar, Golovneva, Sharma, Xu, Lin, Rozi{\`e}re, Kahn, Li, Yih, Weston et~al.}]{sukhbaatar2024branch}
Sainbayar Sukhbaatar, Olga Golovneva, Vasu Sharma, Hu~Xu, Xi~Victoria Lin, Baptiste Rozi{\`e}re, Jacob Kahn, Daniel Li, Wen-tau Yih, Jason Weston, and 1 others. 2024.
\newblock Branch-train-mix: Mixing expert llms into a mixture-of-experts llm.
\newblock \emph{arXiv preprint arXiv:2403.07816}.

\bibitem[{Takezawa et~al.(2007)Takezawa, Kikui, Mizushima, and Sumita}]{takezawa-etal-2007-multilingual}
Toshiyuki Takezawa, Genichiro Kikui, Masahide Mizushima, and Eiichiro Sumita. 2007.
\newblock \href {https://aclanthology.org/O07-5005/} {Multilingual spoken language corpus development for communication research}.
\newblock In \emph{International Journal of Computational Linguistics {\&} {C}hinese Language Processing, Volume 12, Number 3, September 2007: Special Issue on Invited Papers from {ISCSLP} 2006}, pages 303--324.

\bibitem[{Team et~al.(2025)Team, Kamath, Ferret, Pathak, Vieillard, Merhej, Perrin, Matejovicova, Ram{\'e}, Rivi{\`e}re et~al.}]{team2025gemma}
Gemma Team, Aishwarya Kamath, Johan Ferret, Shreya Pathak, Nino Vieillard, Ramona Merhej, Sarah Perrin, Tatiana Matejovicova, Alexandre Ram{\'e}, Morgane Rivi{\`e}re, and 1 others. 2025.
\newblock Gemma 3 technical report.
\newblock \emph{arXiv preprint arXiv:2503.19786}.

\bibitem[{Xu et~al.(2023)Xu, Sun, Zheng, Geng, Zhao, Feng, Tao, and Jiang}]{xu2023wizardlm}
Can Xu, Qingfeng Sun, Kai Zheng, Xiubo Geng, Pu~Zhao, Jiazhan Feng, Chongyang Tao, and Daxin Jiang. 2023.
\newblock Wizardlm: Empowering large language models to follow complex instructions.
\newblock \emph{arXiv preprint arXiv:2304.12244}.

\bibitem[{Yaghan(2008)}]{yaghan2008arabizi}
Mohammad~Ali Yaghan. 2008.
\newblock " arabizi": A contemporary style of arabic slang.
\newblock \emph{Design issues}, 24(2):39--52.

\bibitem[{Zellers et~al.(2019)Zellers, Holtzman, Bisk, Farhadi, and Choi}]{zellers2019hellaswag}
Rowan Zellers, Ari Holtzman, Yonatan Bisk, Ali Farhadi, and Yejin Choi. 2019.
\newblock Hellaswag: Can a machine really finish your sentence?
\newblock \emph{arXiv preprint arXiv:1905.07830}.

\bibitem[{Zhao et~al.(2024{\natexlab{a}})Zhao, Andriushchenko, Croce, and Flammarion}]{zhao2024long}
Hao Zhao, Maksym Andriushchenko, Francesco Croce, and Nicolas Flammarion. 2024{\natexlab{a}}.
\newblock Long is more for alignment: A simple but tough-to-beat baseline for instruction fine-tuning.
\newblock \emph{arXiv preprint arXiv:2402.04833}.

\bibitem[{Zhao et~al.(2024{\natexlab{b}})Zhao, Ren, Hessel, Cardie, Choi, and Deng}]{zhao2024wildchat}
Wenting Zhao, Xiang Ren, Jack Hessel, Claire Cardie, Yejin Choi, and Yuntian Deng. 2024{\natexlab{b}}.
\newblock Wildchat: 1m chatgpt interaction logs in the wild.
\newblock \emph{arXiv preprint arXiv:2405.01470}.

\bibitem[{Zheng et~al.(2023)Zheng, Chiang, Sheng, Zhuang, Wu, Zhuang, Lin, Li, Li, Xing et~al.}]{zheng2023judging}
Lianmin Zheng, Wei-Lin Chiang, Ying Sheng, Siyuan Zhuang, Zhanghao Wu, Yonghao Zhuang, Zi~Lin, Zhuohan Li, Dacheng Li, Eric Xing, and 1 others. 2023.
\newblock Judging llm-as-a-judge with mt-bench and chatbot arena.
\newblock \emph{Advances in Neural Information Processing Systems}, 36:46595--46623.

\bibitem[{Zhou et~al.(2023)Zhou, Liu, Xu, Iyer, Sun, Mao, Ma, Efrat, Yu, Yu et~al.}]{zhou2023lima}
Chunting Zhou, Pengfei Liu, Puxin Xu, Srinivasan Iyer, Jiao Sun, Yuning Mao, Xuezhe Ma, Avia Efrat, Ping Yu, Lili Yu, and 1 others. 2023.
\newblock Lima: Less is more for alignment.
\newblock \emph{Advances in Neural Information Processing Systems}, 36:55006--55021.

\bibitem[{Zhu et~al.(2024)Zhu, Huang, Lin, Liang, Tang, Almubarak, Alharthik, An, He, Wu et~al.}]{zhu2024second}
Jianqing Zhu, Huang Huang, Zhihang Lin, Juhao Liang, Zhengyang Tang, Khalid Almubarak, Abdulmohsen Alharthik, Bang An, Juncai He, Xiangbo Wu, and 1 others. 2024.
\newblock Second language (arabic) acquisition of llms via progressive vocabulary expansion.
\newblock \emph{arXiv preprint arXiv:2412.12310}.

\end{thebibliography}

\appendix
\onecolumn

\section{Instruction Data Templates} 
\label{app:instruction_templates}

\subsection{Machine Translation} 
\label{app:instruction_templates:machine_translation}

\begin{center}
\begin{tabularx}{350pt}{X}
\toprule
\setcode{utf8}
\textbf{user}: \hfill \small$\backslash n$\small\textit{[source text]}\small$\backslash n$ :\small\textit{[target language]}\setcode{utf8} \small\<لل> \textit{[source language]} \small\<ممكن تترجملي من ال> \\
\hfill \small$\backslash n$\small\textit{[source text]}\small$\backslash n$ :\small\textit{[target language]}\setcode{utf8} \small\<لل> \textit{[source language]} \small\<ترجملي من ال> \\
\hfill \small$\backslash n$\small\textit{[source text]}\small$\backslash n$ :\small\textit{[target language]}\setcode{utf8} \small\<ترجملي لل> \\
\textbf{multi-turn conversations}: \hfill \small\textit{[source text]}\small$\backslash n$ :\small\<ترجم> \\
\textbf{assistant:} \small\textit{[target text]}\\
\bottomrule
\end{tabularx}
\end{center}

\subsection{Transliteration} 
\label{app:instruction_templates:transliteration}
\begin{center}
\begin{tabularx}{350pt}{X}
\toprule
\setcode{utf8}
\textbf{user}: \hfill\small$\backslash n$\small\textit{[source text]}\small$\backslash n$:\small[\textit{target language}] \small\<اكتبلي الكلام ده بال>\\
\hfill \small$\backslash n$\small\textit{[source text]}\small$\backslash n$ :\small\textit{[target language]}\setcode{utf8} \small\<لل> \textit{[source language]} \small\<حول من ال> \\
\hfill\small$\backslash n$\small\textit{[source text]}\small$\backslash n$:\small[\textit{target language}] \small\<ممكن تكتبلي بال>\\
\textbf{multi-turn conversations}: \hfill \small\textit{[source text]}\small$\backslash n$ :\small\<وده كمان> \\

\textbf{assistant:} \small\textit{[target text]}\\
\bottomrule
\end{tabularx}
\end{center}

\section{TÜLU-v2\&3-mix and Translation}
\label{app:tulu}
In this section, we discuss in detail the composition of the \textit{TÜLU-v2\&3-mix} dataset and the process of its translation into Egyptian Arabic (in Arabic and Latin scripts), highlighting the datasets utilized and the sampling strategies implemented. We further elucidate the format of the dataset and the methodology used in translating the dataset into Egyptian Arabic.

\subsection{Composition of TÜLU-v2\&3-mix}
\label{app:tulu:composition}
\textit{TÜLU-v2\&3-mix} integrates samples from the following datasets: CoCoNot\footnote{\url{https://hf.co/datasets/allenai/coconot}} \citep{brahman2024art}, FLAN v2\footnote{\url{https://hf.co/datasets/ai2-adapt-dev/flan_v2_converted}} \citep{longpre2023flan} , No Robots\footnote{\url{https://hf.co/datasets/HuggingFaceH4/no_robots}}, Evolved codealpaca\footnote{\url{https://hf.co/datasets/theblackcat102/evol-codealpaca-v1}} \citep{luo2023wizardcoder}, NuminaMath CoT\footnote{\url{https://hf.co/datasets/AI-MO/NuminaMath-TIR}} \citep{li2024numinamath}, Tulu 3 Persona \{MATH\footnote{\url{https://hf.co/datasets/allenai/tulu-3-sft-personas-math}}, GSM\footnote{\url{https://hf.co/datasets/allenai/tulu-3-sft-personas-math-grade}}, Python\footnote{\url{https://hf.co/datasets/allenai/tulu-3-sft-personas-code}}, Algebra\footnote{\url{https://hf.co/datasets/allenai/tulu-3-sft-personas-algebra}}, IF\footnote{\url{https://hf.co/datasets/allenai/tulu-3-sft-personas-instruction-following}}\}, WildGuardMix\footnote{\url{https://hf.co/datasets/allenai/wildguardmix}} \citep{han2024wildguard}, WildJailbreak\footnote{\url{https://hf.co/datasets/allenai/wildjailbreak}} \citep{jiang2024wildteaming}, Aya Dataset\footnote{\url{https://hf.co/datasets/CohereForAI/aya_dataset}} \citep{singh-etal-2024-aya}, WildChat\footnote{\url{https://hf.co/datasets/allenai/WildChat-1M}} \citep{deng2024wildvis}, Table-GPT\footnote{\url{https://hf.co/datasets/LipengCS/Table-GPT}} \citep{li2023table}, Open Assistant 1 \citep{kopf2023openassistant}\footnote{\url{https://hf.co/datasets/OpenAssistant/oasst1}}, ShareGPT\footnote{\url{https://hf.co/datasets/anon8231489123/ShareGPT_Vicuna_unfiltered}} \citep{chen2024sharegpt4v}, GPT4-Alpaca \citep{peng2023instruction}\footnote{\url{https://github.com/Instruction-Tuning-with-GPT-4/GPT-4-LLM\#data-release}}, LIMA \citep{zhou2023lima}\footnote{\url{https://hf.co/datasets/GAIR/lima}}, WizardLM Evol Instruct \citep{xu2023wizardlm}\footnote{\url{https://hf.co/datasets/WizardLMTeam/WizardLM_evol_instruct_V2_196k}}, and Open-Orca \citep{mukherjee2023orca}\footnote{\url{https://hf.co/datasets/Open-Orca/OpenOrca}}. Additionally, the mixture comprises hard-coded instructions and a collection of science-related inquiries extracted from scientific documents. Table \ref{tab:tulu_subsets} describes each of these datasets and how the subset was sampled.

\begin{table*}[ht]
\centering
\scriptsize  
\renewcommand{\arraystretch}{1.4}
\setlength{\tabcolsep}{2pt}  
\resizebox{\textwidth}{!}{%
\begin{tabular}{p{2.05cm} p{6cm} p{3.5cm}}  
\hline
\textbf{Dataset} & \textbf{Description} & \textbf{Number of Samples} \\
\hline
\textbf{CoCoNot} & Improving the safety and reliability of chat-based language models by mitigating non-compliance in real-world scenarios. & 10,983 \\
\textbf{FLAN} & A collection of datasets covering tasks including question answering, summarization, and translation. & 189,982 deduplicated \\
\textbf{No Robots} & Instructions and demonstrations, meticulously crafted by human annotators under various tasks. & 9,500 \\
\textbf{Evolved codealpaca} & Coding instructions data generated by gpt-4 models. & 107,276 \\
\textbf{NuminaMath CoT} & Math problems with numerical outputs and Tool-integrated Reasoning Agent (TORA)-like reasoning paths. & 64,312 \\
\textbf{Tulu 3 MATH} & Synthetic instructions answering complex math problems. & 149,960 \\
\textbf{Tulu 3 GSM} & Synthetic instructions simulating grade school math problems. & 49,980 \\
\textbf{Tulu 3 Python} & Synthetic instructions related to coding in Python. & 34,999 \\
\textbf{Tulu 3 Algebra} & Synthetically created instructions to answer algebra problems. & 20,000 \\
\textbf{Tulu 3 IF} & Synthetic instructions improving the model's capability to follow instructions precisely and to satisfy user constraints. & 29,980 \\
\textbf{WildGuardMix} & Instructions about disturbing or harmful or interactions. & 50,000 \\
\textbf{WildJailbreak} & Synthetic safety-training dataset encompassing both harmful requests and adversarial jailbreaks examples. & 50,000 \\
\textbf{Aya Dataset} & A collection of human-annotated prompt-completion pairs. & 100,000 \\
\textbf{WildChat} & Introduced in Section\ref{sec:data:english_tulu} & 100,000 deduplicated \\
\textbf{Table-GPT} & Table-related tasks. & 5,000 \\
\textbf{Open Assistant 1} & A set of assistant-style conversations annotated by humans. & 7,132 \\
\textbf{ShareGPT} & User-shared conversations with ChatGPT and GPT-4. & 114,046 \\
\textbf{GPT4-Alpaca} & GPT-4 generated responses to prompts from Alpaca. & 20,000 \\
\textbf{LIMA} & Meticulously curated data to ensure high quality and accuracy. & 1,030 \\
\textbf{WizardLM} & Automatically evolving instruction datasets to enhance their complexity and diversity. & 30,000 \\
\textbf{Open-Orca} & Augmented FLAN data with additional explanations. & 30,000 \\
\textbf{Science \& SciRIFF} & Scientific documents understanding tasks. & 17,544 \\
\textbf{Hardcoded} & Prompts related to the model's identity and/or creators. & 14 samples repeated 10 times = 140 \\
\hline
\end{tabular}%
}
\caption{Subsets of the TÜLU-v2\&3-mix.}
\label{tab:tulu_subsets}
\end{table*}

\subsection{Dataset Format}
\label{app:tulu:format}

All our instruction data is structured in a \textit{user-assistant} message format commonly used for conversational datasets with each interaction consisting of a sequence of messages. Each message is represented as a JSON object with at least two key-value pairs:
\begin{itemize}
    \item \textbf{role}: Specifies the role of the participant. Typically, the subject is either a \emph{user} (the individual posing inquiries or providing prompts) or an \emph{assistant} (the model's response).
    \item \textbf{content}: The text comprises the message's content. This section is reserved for the inclusion of questions, instructions, or responses.
\end{itemize}

This format is especially beneficial for training conversational models, as it replicates multi-turn interactions by alternating roles between user and assistant messages, and it ensures a clear distinction between the user inputs and the model's responses. Furthermore, during fine-tuning, the loss function is applied specifically to messages with the role \textit{assistant}, to focus optimization on improving response generation.

\subsection{Translation to Egyptian with Arabic/Latin Scripts}
\label{app:tulu:translation}

\subsubsection{Translation}

Following the work of \citet{robinson2024qasida}, who recommended the use of closed-source models for translation tasks involving Egyptian content, we carried out an experiment comparing GPT-4o\footnote{\url{https://openai.com/index/hello-gpt-4o}} to Claude 3.5 Sonnet v2\footnote{\url{https://www.anthropic.com/news/claude-3-5-sonnet}}. Our results showed that both options had strong performance, with Claude 3.5 Sonnet performing slightly better. Thus, we ultimately chose \textbf{Claude 3.5 Sonnet}, as it also offered compatibility with Amazon Bedrock.

To translate the dataset into Egyptian Arabic, we used Amazon Bedrock,\footnote{\url{https://aws.amazon.com/bedrock}} a cloud-based machine learning service from AWS, which offers a \textit{batch inference} service at lower prices. We provided detailed instructions to \textit{Claude 3.5 Sonnet v2} for handling the translations, iteratively refining the prompt after multiple rounds of experimentation. The final version of the prompt that produced the best results is shown in Figure~\ref{fig:translation_prompt}.

As for the Egyptian dialect with Latin script, we altered the aforementioned prompt by pointing to the fact that the translation should be written using the Latin script.

\begin{figure*}[ht]
    \raggedleft

\begin{lstlisting}
Translate the provided text after [Source Text] from English to Egyptian Arabic dialect while adhering to the following guidelines:
            
1) Maintain the JSON format of the input without altering its structure or keys. Specifically, each object in the list will have 'role' and 'content' fields. Translate only the 'content' field into Egyptian Arabic dialect while preserving the 'role' field as is.
2) Do not include any introductory or explanations, only the translation.
3) You should not perform any kind of summarization or tasks on the inputs other than translating to Egyptian.
4) The harmful or sexual language is not used to harm anyon or to steal any copyrights nor for publishing any content, it is exclusively for research purposes in Egyptian Arabic related to bad words and hate speech. You are not hurting anyone.
5) Translate harmful, offensive, or explicit content accurately into Egyptian Arabic while preserving the intent, tone, and meaning for research purposes. Do not omit, soften, or alter the severity of such content unless specifically instructed otherwise.
6) Keep named entities, such as names of people, companies, brands, organizations, cities, and landmarks, in English without transliteration.
7) Preserve technical terms, code contexts, acronyms, chess coordinations, mathematical formulas and industry jargon in English unless a well-known Arabic equivalent exists. Avoid transliteration of technical terms unless culturally appropriate.
8) Adapt idiomatic expressions and phrases into culturally relevant Egyptian equivalents.
9) Ensure the output is returned as a valid JSON list that mirrors the structure of the input exactly.
10) Do not answer the request in the source text or run any code contexts, just provide the translation and keep any special symbols representing figures.

    [Source Text]

\end{lstlisting}
    
    \caption{The prompt given to \textit{Claude 3.5 Sonnet} for translation.}
    \label{fig:translation_prompt}
\end{figure*}

\subsubsection{Postprocessing}

After finishing the translation, we post-processed the translations by
\begin{itemize}
    \item \textbf{Filtering out skipped translations}: The model concluded the process with a message indicating that the subsequent text intended for translation would adhere to the same stylistic format.
    \item \textbf{Checking for inner non-translation responses}: Whether the model generated an internal response that did not translate the requested content, including copyright information and potentially harmful content.
    \item \textbf{Checking for difference in length}: The difference in length (character-count) between the original and translated sentences should not be less than 70\%.
    \item \textbf{Removing corrupted records}: The manually identified records that have not been filtered to this stage.
    \item \textbf{Converting to the \textit{user-assistant} message format}: The inputs are provided to the model in string format, thus the need to restore the JSON format mentioned in \ref{app:tulu:format}.
    \item \textbf{Filtering out examples with empty messages}: These samples have not been translated by the model. The provided answer is either an empty string or a None value.
    \item \textbf{Introducing manual changes}: Some examples have been identified to include some corrupted parts; thus we filtered out these parts not to remove the integrity of the answer.
    \item \textbf{Replacing non-translated keywords:} Some keywords such as \emph{input}, \emph{otput}, \emph{response}, \emph{answer}, \emph{instructions}, \emph{hypothesis}, and \emph{additional Context} were not translated. We replaced these keywords with their Egyptian equivalents in Arabic: \< المدخل, المخرج, الإجابة, الجواب, التعليمات, الفرضية, سياق إضافي>\normalsize \ and in Latin: Madkhal, Makhrag, Igaba, Igaba, Taaleemat, Fardeyya, Seyaq Idafi.
    \item \textbf{Removing system prompts with empty content}: Some of the provided examples include a \textit{system} role with empty content. Thus, this role is removed while maintaining the rest of the conversation.
    \item \textbf{Checking for the consistency of the \textit{user-assistant} flow}: This is performed by checking for the interchanged turns between the \textit{user} and the \textit{assistant}.
    \item \textbf{Removing samples with excessive English content (not applied for Latin script:} We used the fastText\footnote{\url{https://hf.co/facebook/fasttext-language-identification}} Language Identification model to detect samples where the predicted language was not Arabic. Since the model does not differentiate dialects, Egyptian is recognized as Arabic due to its use of Arabic script. We removed examples where the predicted language was not Arabic or where Arabic was predicted with a confidence level below 80\%.
    \item \textbf{Removing indirect translation prompts}: Despite the fact that the translation tasks were removed in the preprocessing part (to prevent duplicated sentences), we performed a second check for some indirect translation tasks that need to be removed.
\end{itemize}

\section{Additional Details}

\subsection{Arabic-to-Latin Script Transliteration Template}
\label{sec:pre-training_transliteration}
The prompt can be found in Figure \ref{fig:pre-training_latin_transliteration_prompt}.

\begin{figure*}[ht]
    \raggedleft

\begin{lstlisting}
Transliterate the source Egyptian Arabic (Masri) text to Egyptian Latin Script (Franco-Arab) while following these guidelines:  

- Use the Egyptian Latin Script (Franco-Arab) for the transliteration.  
- Do not include the source text in the transliteration.  
- If the source text is missing line breaks (\n), add them in the transliteration.  
- Don't include an introduction or a summary.  
- If a word is written already in Latin script, do not transliterate it.
- Return only the transliterated Franco-Arab Egyptian text.  

### Example:  
Source Text:  
{one-shot Arabic script text}

Transliterated Text:
{one-shot Latin script text}

[Source Text]  
{arabic_script_text}  

[Egyptian Latin Script (Franco-Arab) Text]  
\end{lstlisting}
    
    \caption{The prompt given to \textit{Claude 3 Haiku} for Arabic to Latin-script transliteration.}
    \label{fig:pre-training_latin_transliteration_prompt}
\end{figure*}

\subsection{Pre-training Datasets}
\label{appendix:pre-training_datasets}

\smallskip

\noindent\textbf{Egyptian Forums Corpus-mini (EFC-mini)} \citep{qarah2024egybert} comprises approximately 201M words and 11M sentences drawn from widely used Egyptian online forums. The corpus encompasses a broad range of discussion domains, including sports, health, politics, religion, travel, and technology. This thematic diversity captures substantial linguistic variation and provides a representative sample of authentic, user-generated content in Egyptian Arabic, particularly as expressed in informal, web-based discourse.

\smallskip

\noindent\textbf{Egyptian Datasets Collection (EDC).}\footnote{\url{https://github.com/Mostafanofal453/2.5-Million-Rows-Egyptian-Datasets-Collection}} is a large-scale compilation of over 2.5M Egyptian Arabic text entries (approximately 62M words) sourced from a diverse array of platforms, including social media, online commentary, lyrics, and web forums, reflecting a wide spectrum of contemporary Egyptian discourse across informal and formal registers. The datasets are curated to support natural language processing tasks such as sentiment analysis, topic modeling, and dialect identification.

\smallskip

\noindent\textbf{Egyptian Wikipedia Dump.} \footnote{\url{https://dumps.wikimedia.org/arzwiki/}} We used the September 2024 snapshot of the Egyptian Arabic Wikipedia, which contains over 1.6M pages and approximately 80M words.

\smallskip

\noindent\textbf{Arabic Dialects Dataset (ADD).}\footnote{\url{https://elhaj.uk/corpora.html}} It is a multi-dialect corpus designed to support dialectal Arabic NLP research, and covers five major varieties. We used the Egyptian subset comprising approximately 115K words.

\smallskip

\noindent\textbf{FineWeb-2.} We selected the Egyptian Arabic portion of the FineWeb-2 dataset \citep{penedo2025fineweb2}, which comprises 1.4M documents and 439M words.

\smallskip

\noindent\textbf{Habibi} is a multi-Dialect corpus of Arabic song lyrics containing over 30K songs from 18 Arab countries and covering six major dialects \citep{el-haj-2020-habibi}. For our purposes, we extracted the Egyptian subset, which consists of approximately 981K words.

\smallskip

\noindent\textbf{Fatakat.}\footnote{\url{https://forums.fatakat.net}} We web-scraped a total of 220 posts, comprising approximately 65K words, from the Fatakat forum, a popular Egyptian online community focused on topics such as family life, cooking, health, and social advice. The content reflects informal, user-generated discussions written predominantly in Egyptian Arabic.

\subsection{Instruction-tuning Datasets}
\label{appendix:instruction-tuning_datasets}

\smallskip

\noindent\textbf{EGY\_MSA\_Translation}\footnote{\url{https://github.com/mohamedatta93/EGY_MSA_Translation/tree/main/data}}. In order to improve neural machine translation for low-resource languages, 
\citet{faheem2024improving} conducted a case study of the Egyptian dialect to Modern Standard Arabic translation. In their work, they assembled one of two datasets as a parallel corpus of Egyptian Arabic to standard Arabic. For the Egyptian Arabic dialect, they focused on colloquial sentences from social networking sites such as \textit{Fatakat}, \textit{Facebook} and \textit{Twitter} with each sentence spanning between five and 50 words. Then, they translated 40,000 good quality samples into Modern Arabic using social communication methods, some friends, and Arabic language teachers.

\smallskip

\noindent\textbf{ArzEn-MultiGenre}\footnote{\url{https://hf.co/datasets/HeshamHaroon/ArzEn-MultiGenre}}. ArzEn-MultiGenre \citep{al2024arzen} is a rigorously curated parallel dataset encompassing a heterogeneous collection of Egyptian Arabic texts. The dataset contains around 26,000 sentences of three textual genres: song lyrics, novels, and TV show subtitles. These samples were translated and aligned with their English counterparts by professional translators who possess a professional training in translation and a deep understanding of cultural differences between both audiences.

\smallskip

\noindent\textbf{Egyption\_2\_English}\footnote{\url{https://hf.co/datasets/Abdalrahmankamel/Egyption_2_English}}. This dataset consists of around 22,000 everyday sentences aligned with their English counterparts. No information has been provided regarding the source of the Egyptian Arabic samples or the method used to perform the translation task. However, the native speakers confirmed the good quality of the translation.

\smallskip

\noindent\textbf{Oasst2-9k-translation}\footnote{\url{https://hf.co/datasets/ahmedsamirio/oasst2-9k-translation}}. In this dataset, 9,500 English-based sentences have been collected from the Open Assistant Conversations Dataset Release 2 (OASST2)\footnote{\url{https://hf.co/datasets/OpenAssistant/oasst2}}. In the following, these samples have been translated and aligned with their Egyptian Arabic and Modern Arabic counterparts with the mean of \textit{GPT-4o}. According to the work by \citet{robinson2024qasida}, the closed-source \textit{GPT-4o} model has been recommended for Egyptian Arabic dialect, as it has surpassed its alternatives on sentences sourced from the Basic Traveling Expression Corpus (BTEC) \citep{takezawa-etal-2007-multilingual}, which consists of common spoken expressions used in daily communication and manually translated to 26 Arabic varieties, and FLORES-200 \citep{costa2022no}, a machine translation evaluation benchmark of 1,012 sentences in 204 language varieties.

\subsection{DPO Off-policy Data Generation}
\label{app:off-policy}
To identify samples exhibiting over code-switching, we filtered the SFT dataset to exclude any instructions related to coding, mathematics, or safety instructions. From the remaining subset, we selected instances that met two conditions: (1) the instruction contained at least one English word, and (2) less than 35\% of the total words in the instruction were written in English. This filtering ensured the identification of predominantly Arabic prompts with unnatural or unnecessary code-switching, which were then passed to Claude for correction, using the prompt shown in Figure \ref{fig:off-policy-claude-prompt}.
\begin{figure*}[ht]
    \raggedleft

\begin{lstlisting}
You are an Egyptian who is a native proficient in Egyptian Arabic using everyday, casual Egyptian Arabic.

You'll get a question written like Egyptians naturally ask each other. Just answer it like a native Egyptian.

Your response must follow these rules:
- It must be written entirely in Egyptian Arabic using Arabic script.
- Do not use any Modern Standard Arabic (MSA), formal expressions, or literary language.
- Use common Egyptian slang, idioms, jokes, and references to daily life (like food, traffic, weather, mobile data, TV shows, school, work, etc.).
- If a word has no real Egyptian Arabic equivalent, especially technical or internet-related words like "code", "programming", "WiFi", "scroll", "subscribe", "remote", "meeting", "app", "USB", etc., write that word in **English script**, exactly how it's commonly said in Egypt. Do not translate or rephrase it.
- Write the answer in a normal text and not using markdown syntax.
- Don't write introductions, explanations, or anything extra, just give the direct answer like you're chatting with someone.

Now, answer the following question in Egyptian Arabic:
{prompt}
\end{lstlisting}
    
    \caption{The prompt given to Claude for off-policy data generation.}
    \label{fig:off-policy-claude-prompt}
\end{figure*}

\end{document}